\documentclass[sigconf]{acmart}
\AtBeginDocument{%
  \providecommand\BibTeX{{%
    \normalfont B\kern-0.5em{\scshape i\kern-0.25em b}\kern-0.8em\TeX}}}

\usepackage{psfrag, float}
\usepackage{algorithm}
\usepackage{algorithmic}
\usepackage{svg}
\usepackage{pdfpages}
\usepackage{microtype}
\usepackage{booktabs} % for professional tables
\usepackage{subfig}
\usepackage{graphicx} % DO NOT CHANGE THIS
\usepackage{multirow}
\usepackage{color}

\providecommand{\keywords}[1]{\textbf{\textit{Index terms---}} #1}

\def\eqref#1{eq.~(\ref{#1})}
\def\1{\bm{1}}

\DeclareMathAlphabet{\mathsfit}{\encodingdefault}{\sfdefault}{m}{sl}
\SetMathAlphabet{\mathsfit}{bold}{\encodingdefault}{\sfdefault}{bx}{n}

\newcommand{\eg}{\textit{e.g.}}

\definecolor{orange}{RGB}{255,107,0}

\copyrightyear{2021}
\acmYear{2021}
\setcopyright{rightsretained}
\acmConference[ACM CHIL '21]{ACM Conference on Health, Inference, and Learning}{April 8--10, 2021}{Virtual Event, USA}
\acmBooktitle{ACM Conference on Health, Inference, and Learning (ACM CHIL '21), April 8--10, 2021, Virtual Event, USA}\acmDOI{10.1145/3450439.3451865}
\acmISBN{978-1-4503-8359-2/21/04}

\acmSubmissionID{73}

\begin{document}

%%
%% The "title" command has an optional parameter,
%% allowing the author to define a "short title" to be used in page headers.
\title{iGOS++: Integrated Gradient Optimized Saliency by Bilateral Perturbations}

%%
%% The "author" command and its associated commands are used to define
%% the authors and their affiliations.
%% Of note is the shared affiliation of the first two authors, and the
%% "authornote" and "authornotemark" commands
%% used to denote shared contribution to the research.

\author{Saeed Khorram}
\authornote{Equal contribution.}
\affiliation{%
  \institution{Collaborative Robotics and Intelligent Systems (CoRIS) Institute, Oregon State University}
  \country{USA}
}
\email{khorrams@oregonstate.edu}

\author{Tyler Lawson\footnotemark[1]}
\affiliation{%
  \institution{Collaborative Robotics and Intelligent Systems (CoRIS) Institute, Oregon State University}
  \country{USA}
}
\email{lawsont@oregonstate.edu}

\author{Li Fuxin}
\affiliation{%
  \institution{Collaborative Robotics and Intelligent Systems (CoRIS) Institute, Oregon State University}
  \country{USA}
}
\email{lif@oregonstate.edu}

%%
%% By default, the full list of authors will be used in the page
%% headers. Often, this list is too long, and will overlap
%% other information printed in the page headers. This command allows
%% the author to define a more concise list
%% of authors' names for this purpose.

% \renewcommand{\shortauthors}{Trovato and Tobin, et al.}

%%
%% The abstract is a short summary of the work to be presented in the
%% article.
%-------------------------------------------------------------------------
\begin{abstract}
The black-box nature of the deep networks makes the explanation for "why" they make certain predictions extremely challenging. Saliency maps are one of the most widely-used local explanation tools to alleviate this problem. One of the primary approaches for generating saliency maps is by optimizing for a mask over the input dimensions so that the output of the network for a given class is influenced the most. However, prior work only studies such influence by removing evidence from the input. In this paper, we present iGOS++, a framework to generate saliency maps for black-box networks by considering both removal and preservation of evidence. Additionally, we introduce the bilateral total variation term to the optimization that improves the continuity of the saliency map especially under high resolution and with thin object parts. We validate the capabilities of iGOS++ by extensive experiments and comparison against state-of-the-art saliency map methods. Our results show significant improvement in locating salient regions that are directly interpretable by humans. Besides, we showcased the capabilities of our method, iGOS++, in a real-world application of AI on medical data: the task of classifying COVID-19 cases from x-ray images. To our surprise, we discovered that sometimes the classifier is overfitted to the text characters printed on the x-ray images when performing classification rather than focusing on the evidence in the lungs. Fixing this overfitting issue by data cleansing significantly improved the precision and recall of the classifier.

\end{abstract}

%%
%% The code below is generated by the tool at http://dl.acm.org/ccs.cfm.
%% Please copy and paste the code instead of the example below.
%%


%%
%% Keywords. The author(s) should pick words that accurately describe
%% the work being presented. Separate the keywords with commas.

%% A "teaser" image appears between the author and affiliation
%% information and the body of the document, and typically spans the
%% page.
% \begin{teaserfigure}
%   \includegraphics[width=\textwidth]{sampleteaser}
%   \caption{Seattle Mariners at Spring Training, 2010.}
%   \Description{Enjoying the baseball game from the third-base
%   seats. Ichiro Suzuki preparing to bat.}
%   \label{fig:teaser}
% \end{teaserfigure}

%%
%% This command processes the author and affiliation and title
%% information and builds the first part of the formatted document.
\maketitle

%-------------------------------------------------------------------------
\section{Introduction} \label{sec:intro}
As deep networks achieve excellent performance in many tasks, more and more people want to open these black boxes to understand how they make their decisions under the hood. Especially, explaining deep classifiers can help to potentially ``debug" them to understand how they make mistakes, and fix those mistakes e.g. by additional data preprocessing. This is increasingly important as deep learning is starting to be used in critical decision-making scenarios such as autonomous driving and medical diagnosis.

Saliency map or heatmap visualization is a fundamental tool for explaining convolutional networks (CNNs). It is mostly used to directly explain classification decisions, but approaches that explain intermediate network nodes or forming new concepts would depend on them as well. In earlier days, heatmap visualizations were mostly based on computing gradient variants of the network with respect to the input~\cite{Gradcam17,zhang16excitationBP,SimonyanVZ13}. However, due to the highly nonlinear nature of the CNNs, those one-step gradients only account for infinitesimal changes in the function values and do not necessarily correlate to the features that CNNs actually use for decision making \cite{adebayo2018sanity}. This has caused some of the earlier works to be disillusioned on heatmap research. The most successful of the gradient-based approaches, Grad-CAM~\cite{Gradcam17}, partially avoids this issue by not backpropagating into the convolutional layers, hence correlating reasonably well with CNN classification. However, its heatmaps are quite low-resolution since it only works at the final layers where the images are already low-resolution.

\begin{figure*}[bt]
\begin{center}

\includegraphics[width=0.99\linewidth]{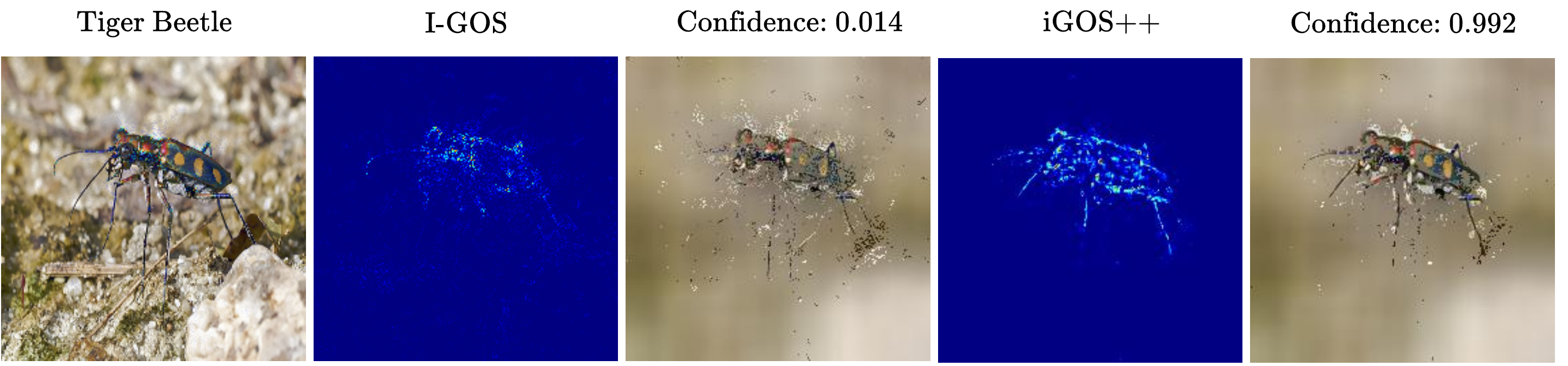}   % uncomment if two rows are too much

\end{center}
% \vspace{-.15in}
\caption{\small Comparison of heatmaps generated from \cite{IGOS} (I-GOS) and our proposed approach (iGOS++) for a \textit{tiger beetle} image. I-GOS focuses only on \textit{breaking} existing evidence (e.g. legs of the beetle), hence generated a mask that is highly scattered under high resolution (2nd image from left), and prediction confidence with the masked area (middle image) on the tiger beetle category is very low. However, with the same amount of pixels ($6\%$)  from the proposed iGOS++ heatmap, the network has $99.2\%$ confidence (5th image from left)} 
\label{fig:tiger_beetle}
 \vskip -0.15in
\end{figure*}

Another group of approaches optimizes for a small mask so that CNNs can no longer classify masked images~\cite{ClassicMask,IGOS}. These approaches provably correlate with the CNN classification, since it can usually be shown that the CNN would no longer predict the original category once the image is masked. However, as we experiment with those approaches, we find that sometimes it is much simpler to ``break" the important features CNNs are using to classify, without necessarily capturing all the important features. An intuitive example of this is an object with its important parts being long and thin, as shown in Fig.~\ref{fig:tiger_beetle}. The mask only needs to cover a small number of points to ``break" the legs of the \textit{tiger beetle}, making it disconnected and making the CNN no longer capable of classification. However,  those areas, when revealed to the CNN, do not necessarily contain enough information (e.g. the complete legs) for CNNs to come to a correct classification of the image. As a result, those methods do not perform well especially at higher resolutions, as their masks become disconnected and ``adversarial", focusing only on reducing the CNN prediction rather than locating all the informative areas. Usually, the capability of CNN classifying the image to its original category based on the masked parts drops noticeably when moving from $14\times 14$ heatmap resolution to e.g. $224 \times 224$~\cite{IGOS}.

In this paper, we propose novel improvements to address this gap. The first novelty is to optimize for an additional insertion loss which aims for the CNN to correctly classify the object even if only a small informative part of the image is revealed. Instead of just adding a simple loss term, we found that separating the deletion mask and insertion mask during the optimization process helps the visualization performance. Our second novelty is to propose a new bilateral total variation term that makes the mask smooth on image areas with similar color, alleviating heatmap diffusion at higher resolutions. A combination of these techniques enables us to obtain heatmap visualizations with significantly better performance, especially when the heatmaps are generated at a high resolution -- which are more prone to adversarial solutions if the masks are derived solely on deletion loss. In the tiger-beetle example (Fig.\ref{fig:tiger_beetle}), it can be seen that iGOS++ is capable of capturing entire legs of the beetle hence capable of generating a high-confidence prediction even with only $6\%$ of the pixels in the original image. We evaluate iGOS++ through extensive experiments and compare our results against the state-of-the-art baselines on non-medical datasets --- ImageNet and FashionMNIST as standard benchmarks for evaluation of saliency maps.

{
We utilized iGOS++ in a real-world problem of detecting COVID-19 patients based on chest x-ray imaging. Interestingly, we found that in some cases, the classifier is overfitted to characters printed on the x-ray images --- which clearly should not be related to the underlying indicators of the disease. This illustrates one of the major problems of the current deep network models: their \textit{lack of interpretability}; These black box classifiers can overfit to class priors not expected by humans, which could lead to poor generalization and arbitrary decisions, particularly in high-stake tasks such as in medical diagnosis. Once we pre-processed all images to remove written characters, meaningful performance improvements on the COVID detection task were observed. This shows the utility of the heatmap visualization algorithms in realistic tasks to open up the black box and reveal some of the biases the classifiers may have learned. 
}

%We believe this paper marks an important step towards a better heatmap visualization which will in turn help other explanation tasks in the future.

% \vspace{-0.15in}

%-------------------------------------------------------------------------
\section{Related Work} \label{sec:related_work}
% \vspace{-0.1in}
%This paper is aims to improve upon the algorithm described in \cite{IGOS}.
%Much of the work is related and like \cite{IGOS} we split the solutions into two general categories: one-step back-propagation based approaches and perturbation based approaches.
We only review related work in heatmap visualizations rather than the broader problem of explaining deep networks. Most visualization approaches can be categorized into gradient-based approaches and perturbation-based approaches.

\textbf{Gradient-based approaches} for generating saliency maps commonly use different backpropagation heuristics to derive the sensitivity of the output score with respect to the input.
Deconv and Saliency Maps~\cite{MatDeconv,SimonyanVZ13} attach special deconvolutional layers to the convolutional layers.
Guided-backprop~\cite{JTguided2015} works in a similar, yet different, method to ~\cite{MatDeconv}, masking values based on negative gradients.
\cite{InputGradient} multiplies the gradient with the image RGB values.
\cite{IntegratedGradient} proposes integrated gradients, which compute multiple gradients along a straight line in the image space and average them.
\cite{LRP15} computes the relevance across different layers and assigns an importance to each pixel that is used to create a saliency map.
\cite{zhang16excitationBP} uses a winner take all probabilistic model to send top-down signals through the network and generate probabilities based on the weights that are used to create the saliencies.
Grad-CAM \cite{Gradcam17} is the most popular visualization method in this category, it  generalizes the existing class activation method~\cite{zhou2016learning} to work on any CNN-based neural network and maps class scores back to the previous convolution layer.
% Unclear what this means, if we cite it we need a better sentence
% SMOE uses the forward-pass activations at the end of each scale, just before down sampling, to create a saliency map.
% Because this is the forward pass and doesn't involve the final output layer, it doesn't contain any class specific information and can't produce a class specific saliency map.
% A class specific saliency map can be obtained by multiplying the SMOE scale saliency map with a GradCAM map.
%\cite{SMOE} is another method that works with a single pass.
%By itself, it can create a saliency map in a single forward pass, but is not class specific. A class specific saliency map can be obtained by combining this saliency map with CAM.

However, gradients reflect \emph{one-step} infinitesimal changes in the input, which do not necessarily correspond to a direction in which the output score from a deep network would drop significantly --- particularly for deep networks that are highly non-linear functions. In addition, some of these saliency maps were shown to be completely or somewhat independent of the category, only showing strong image edges~\cite{adebayo2018sanity}.

%, i.e. %$\nabla f_c(I)$.
%In a neighborhood of a data sample $I_0$ and with the assumption of a smooth $f$, $\nabla f_c(I)$ can locally be approximated using the first-order Taylor expansion,

%{\small 
%    \begin{align} 
%        \nabla f_c(I) \approx \nabla \big( f_c(I_0) + \langle \nabla f_c(I_0), I-I_0 \rangle \big) = \nabla f_c(I_0), 
%    \end{align}\label{eq:taylor}
%}

%\noindent $\nabla f_c(I_0)$ contains the infinitesimal changes in the input $I_0$ that are most relevant to the output score.

%At the most basic level, a one-step back-propagation method computes a single gradient with respect to the image and uses a normalized version of this as a saliency map.
%This has been improved on since by .

\textbf{Perturbation-based approaches} work by modifying the input in some way, e.g. masking, and testing how the output of the network changes.
\cite{Occlude15} is a method that iteratively removes sections of the image using the gradient until the image contains only the information needed for classification of the target class.
In \cite{Dabkowski2017}, a new network is trained to generate saliency maps.
%The saliency map generation is very fast, but requires time and effort to train. 
RISE \cite{2018RISE} and LIME \cite{ribeiro2016should} are similar perturbation based methods that treat the model as a black-box, and thus do not use gradients at all.
They both involve randomly perturbing the image.
RISE weighs all of the random masks by the model's confidence and combines them, taking into account each pixels distribution in the random masks.
In LIME, the random masks are used to fit a linear model to the black-box model in the local space around the image.
The distance from the original image is used in the loss function and the final weights of the linear model are used to generate an explanation for the image.
LIME has a reliance on super-pixels to avoid adversarial masks.
Some methods utilize optimization with multiple iterations to generate a heatmap visualization~\cite{ClassicMask,IGOS}. Here the main challenge is the highly non-convex nature of the optimization problem.
\cite{ClassicMask} optimizes a mask to reduce the prediction confidence of a target class.
Following \cite{ClassicMask}, \cite{fong2019understanding} uses a fixed-area binary mask that maximally effects the output. This is advantageous as it mitigates the balancing issue in the original mask optimization.
\cite{IGOS} is the most related to our work.
It combines the algorithms in \cite{ClassicMask} and \cite{IntegratedGradient}, by utilizing integrated gradient to optimize the mask.
This is shown to significantly improve the performance of the optimization.

% \vspace{-0.15in}

%-------------------------------------------------------------------------
\section{Bilaterally-Optimized Perturbations} \label{sec:method}
% \vspace{-0.15in}
%----------------------------------
\subsection{Background}\label{sec:bg}

%----------------------------------
\subsubsection{Heatmap Visualizations by Optimization} \label{sec:saliency_perturb}
% \footnote{"In this script, Saliency map" terminology in not limited to the work of \cite{SimonyanVZ13}.}}

% The non-transparent learning process in the deep network has characterised them as \emph{black-box} maps to humans. This obstructs the path to develop \emph{global} explanatory for the deep networks. Nevertheless, 
%In order to obtain \emph{local} explanations for the predictions of a black-box network $f: \mathcal{X} \rightarrow \mathcal{Y}$, tools such as \emph{saliency maps} are utilized \cite{SimonyanVZ13,ClassicMask}. 
%Saliency maps provide an importance map over the individual input dimensions for a given input $x \in \mathcal{X}$ and target $y \in \mathcal{Y}$. 
We consider the well-known image classification task, where a black-box network $f$ predicts a score $f_c(I_0)$ on class $c$ for input image $I_0$. Let $\odot$ denote the Hadamard product.

%% Use of e.g. here is awkward. There's no complete sentence without it. Maybe "Prior work optimizes for the deletion task, e.g. masking the image so that it has low predictive confidence for class c."
The idea of optimization-based heatmap visualization is to locate the regions in the input image $I_0$ that are most important for the network $f$ in outputting $f_c(I_0)$. These local perturbations can be formulated by the inner-product of a real-valued \textit{mask} $M$ to the image $f(I_0 \odot M)$ \cite{ClassicMask}. Prior work optimizes for the \textit{deletion task}, namely masking the image so that it has \eg{ low predictive confidence for class $c$}. Afterward, one can visualize the mask to find the salient regions that \emph{caused}  the output confidence to decrease. Mathematically:

%To explain the output of a black-box model $f$ by optimizing for a mask $M$, there are two primary frameworks: \emph{first}, one can observe the change in the output score $f_c(I)$ by \emph{deleting} evidence from different dimensions of the input $I_0$ --- called \textit{deletion} mask throughout the rest of this paper. The work of \cite{ClassicMask, IGOS} are of this kind which the mask $M$ is calculated by the following minimization problem,
% This should be argmin?
{\small
    \begin{align} 
        \min_{M} \; & F_c(I_0, M) = f_c\big(\Phi(I_0,\tilde{I}_0, M)\big) + g(M), \notag\\
        {\rm s.t.} \quad & g(M) = \lambda_1 ||{\bf 1}-M||_1 +\lambda_2 \text{TV}(M),  \label{eq:igos}\\
        & \Phi(I_0,\tilde{I}_0, M) = I_0 \odot  M + \tilde{I}_0 \odot ({\bf 1}-M), \quad {\bf 0} \leq M \leq {\bf 1}, \notag
    \end{align}
}

\noindent where $\tilde{I}_0$ is a \emph{baseline} image with near zero evidence about the target class $c$, $f_c(\tilde{I}_0) \approx \min_{I}f_c(I)$. It is often chosen to be a constant, white noise, or a blurred version of the image $I_0$ \cite{ClassicMask}. The masking operator $\Phi(I_0,\tilde{I}_0, M)$ uses a weighted version between $I_0$ and $\tilde{I}_0$ to block the influence of certain pixels, decided by the mask $M$ values. The aim of optimization problem (\ref{eq:igos}) is to locate a small and smooth mask $M$ which identifies the regions that are most informative to the black-box $f$ by maximally reducing the output score (predictive confidence on class $c$) $f_c\big(\Phi(I_0,\tilde{I}_0, M)\big) \ll f_c(I_0)$. The regularization term $g(M)$ encourages the mask to be small, by penalizing the magnitude of $M$ with coefficient $\lambda_1$, as well as to be smooth, by penalizing the total-variation (TV) in $M$ \cite{ClassicMask} with coefficient $\lambda_2$.

%\emph{Second}, as a complement to the first framework, one can \emph{retain} a minimal set of evidence across dimensions of the input $I_0$ so that output score is maximally preserved $f_c\big(\Phi(I_0,\tilde{I}_0, M)\big) \ge f_c(I_0)$ --- called \textit{insertion} mask throughout the rest of this paper. The advantages of this framework are illustrated more in section \ref{sec:igos++}.

% \vspace{-0.1in}
%----------------------------------
\subsubsection{Integrated Gradient}\label{sec:ig}

Eqn. (\ref{eq:igos}) is a complicated non-convex optimization problem. \cite{ClassicMask} optimizes the mask by gradient descent. However, this is slow and can take hundreds of iterations to converge. In addition, gradient descent can converge to a local optimum and is not able to jump out of it. \cite{IGOS} has alleviated this issue by using \emph{Integrated-gradient} (IG) \cite{IntegratedGradient} rather than conventional gradient as the descent direction for solving the mask optimization. The IG of $f_c(M)$ with respect to $M$ can be formulated as follows,
% with the computational trade-off of calculating the gradient over multiple steps. 
{\small
    \begin{align}
        \nabla_{I_0}^{IG} f_c(M) =\frac{1}{S} \sum_{s=1}^{S} \frac{ \partial f_c\left(\Phi\big(I_0,\tilde{I}_0, \frac{s}{S}M\big)\right)}{\partial M},\label{eq:ig_fc}
    \end{align}
}
\noindent where it accumulates the gradients along the straight-line path from the perturbed image $\Phi\big(I_0,\tilde{I}_0, M\big)$ towards the baseline $\tilde{I_0}$, which approximately solves the global optimum to the unconstrained problem eq. (\ref{eq:igos}). Equivalently, IG can be thought of as performing gradient descent to simultaneously optimize the performance of multiple masks:

\begin{equation}
    \min_M \frac{1}{S} \sum_{s=1}^S f_c\left(\Phi\left(I_0, \tilde{I}_0, \frac{s}{S}M \right)\right) + g(M)
    \label{eq:ig_real}
\end{equation}
which makes it a proper optimization algorithm. Practically, \cite{IGOS} has shown that it improves the optimization performance of eq. (\ref{eq:igos}) as well.

% \vspace{-0.1in}
%----------------------------------
\subsection{Bilateral Minimal Evidence}\label{sec:igos++}
As shown in Fig.~\ref{fig:tiger_beetle} and as discussed in the introduction: the capability of destroying a CNN feature does not by itself fully explain the features a CNN used. In this section we propose I-GOS++, which improves upon \cite{IGOS} by also optimizing for the \textit{insertion task}.
This would make the CNN predict the original class given information from only a \textit{small} and \textit{smooth} area. 
We believe that the deletion task and the \textit{insertion task} are complementary and need to be considered simultaneously, since both of them contain important information that introduces a novel "look" into the network behavior by perturbing the input. Besides, considering one task alone is prone to reach \emph{adversarial} solutions \cite{szegedy2013}, particularly when removing evidence. 
However, it would be unlikely to find adversarial solutions that can satisfy both aforementioned criteria. 
% Further, as stated in \ref{sec:saliency_perturb}, the two frameworks are complementary, and each introduces a novel "look" into the networks behavior when perturbing the input. 
% \ref{fig} <A FIGURE CMP INSETION AND DELETION WITHOUT REGULARIZATION>.\\

A na\"{i}ve approach to implement this would be to add an insertion loss $- f_c\big(\Phi(I_0,\tilde{I}_0, 1 - M)\big)$ directly to eq. (\ref{eq:igos}). %However,  we \textbf{did not} found this approach to show noticeable improvement over the deletion optimization alone \cite{ClassicMask, IGOS}. We speculate that due to the high complexity of the optimization space, simultaneously incorporating the gradients from the insertion loss and the deletion one confuses the optimizer and potentially creates more local optima and saddle points, making the optimization problem hard to solve. 
However, empirically we found that optimizing  separate masks performed better than the direct approach. Our method aims to optimize separate deletion and insertion masks over the input with the constraint that the product of the two also satisfies both the deletion and insertion criteria for a target class $c$, i.e. deletion of the evidence from the input $I_0$ drastically reduces the output score while retaining the same evidence preserves the initial output score $f_c(I_0)$. Formally, we solve the optimization problem with $3$ masks:

{\small
    \begin{align}
        \min_{M =(M_x, M_y)} \; & F_c(I_0, M) =  f_c\big(\Phi(I_0,\tilde{I}_0, M_x)\big) \notag \\
        & - f_c\big(\Phi(I_0,\tilde{I}_0, 1 - M_y)\big)+f_c\big(\Phi(I_0,\tilde{I}_0, M_{xy})\big) \notag \\
        & -f_c\big(\Phi(I_0,\tilde{I}_0, 1 - M_{xy})\big) + g(M_{xy}) \label{eq:igos++}\\
        {\rm subject~to} \quad & g(M_{xy}) = \lambda_1 ||{\bf 1}-M_{xy}||_1 +\lambda_2 \text{BTV}(M_{xy}); \notag \\
        & \quad M_{xy} = M_x \odot M_y; \quad {\bf 0}\leq M_x, M_y \leq {\bf 1} \notag 
    \end{align}
}

\noindent where $M_x$ is the deletion mask, $M_y$ is the insertion mask, and their dot product, $M_{xy}$, is taken as the final solution to the above optimization problem. The BTV term stands for \textit{Bilateral Total Variation} which is explained later in section \ref{sec:btv}.

In the above formulation eq. (\ref{eq:igos++}), the resolution of the masks is flexible. If the mask has lower resolutions than the original input $I_0$, it is first \textit{up-sampled} to the input size using bi-linear interpolation, and then is applied over the image. The choice of the mask resolution depends on the application and the amount of detail desired in the output mask. Commonly, lower resolution masks e.g. $14 \times 14$ tend to generate more coarse and smooth saliency maps, while higher resolution masks, e.g. $224 \times 224$, generate more detailed and scattered ones. Lower resolution masks also have the advantage of being more robust against adversarial solutions \cite{IGOS, szegedy2013}.  In addition, one can add regularization terms on the individual masks $M_x$ and $M_y$, but we have found regularization only on the final mask $M_{xy}$ to be sufficient.

Note that the integrated gradient for the insertion mask $-f_c(\Phi(I_0, $ $\tilde{I}_0, 1 - M))$ is slightly different from Eq. (\ref{eq:ig_fc}) as it calculates the negative IG along the straight-line path from the image perturbed using the inverse mask ${1} - M$, i.e., $\Phi\big(I_0,\tilde{I}_0, {1} - M\big)$, toward the original image $I_0$,

{\small
    \begin{align}
         \nabla_{I_0}^{IG} h_c(M) = - \nabla_{I_0}^{IG} f_c(1-M)  = - \frac{1}{S} \sum_{s=1}^{S} \frac{ \partial f_c\left(\Phi\big(\tilde{I}_0, I_0, \frac{s}{S} M\big)\right)}{\partial M}. \label{eq:ig_hc}
    \end{align}
}

%Building upon the work of \cite{IGOS}, 
We use IG to substitute the conventional gradient for the partial objective $f_c(.)$ in eq. (\ref{eq:igos++}) and the simple gradient for the convex regularization terms $g(\cdot)$. The following is the total-gradient (TG) for each iteration of the mask update:

{\small
    \begin{align}
        {TG}(M) =  & \; \nabla_{I_0}^{IG} f_c(M_x) + \nabla_{I_0}^{IG} h_c(M_y) + \nabla_{I_0}^{IG} f_c(M_{xy}) \\ \notag
        & + \nabla_{I_0}^{IG} h_c(M_{xy}) + \nabla g(M_{xy}). \label{eq:tg}
    \end{align}
}

TG contains separate integrated gradients w.r.t the deletion and insertion masks ($M_x$ and $M_y$). $\nabla^{IG}_{I_0}$ is indicative of the direction to the unconstrained problem eq. (\ref{eq:igos++}) while $\nabla g(M_{xy})$ regularizes the gradients toward a local and smooth mask and discards unrelated information. Moreover, to make the masks to generalize better and be less dependent over the individual dimensions, we add noise to the perturbed image \cite{ClassicMask,IGOS} during each step of the IG calculation in $\nabla_{I_0}^{IG} f_c(M_x) + \nabla_{I_0}^{IG} h_c(M_y)$.% <WHY NOT IN THE M_{XY} MASKS>

\subsubsection{Bilateral Total-Variance}\label{sec:btv}

% One of the primary challenges in generating saliency maps by perturbation \cite{ClassicMask} is to avoid adversarial examples. The optimization problem \ref{eq:igos} (without the TV penalty) is very similar to finding adversarial examples \cite{szegedy2013,goodfellow2014}. Particularly in high resolutions ($224 \times 224$), small perturbations can simply lift the image form its original manifold, i.e., the optimization is prone to find adversarial solutions. To avoid such, following techniques are utilized: \textit{First}, we add noise to the perturbed image when calculating the IG. This follows the similar rational as in \cite{smilkov2017smoothgrad, ClassicMask}, making the mask to generalize better over the input dimensions. \textit{Second}, we upsample the mask from a lower resolution (e.g. $28 \times 28$) to the original input image size using bilinear interpolation. This way, the sharp gradients indispensable for adversarial examples would be avoided. This also results in a smooth mask which is more human understandable.

To further alleviate the problem of scattered heatmaps shown in Fig.~\ref{fig:tiger_beetle}, we introduce a new variation of the TV loss \cite{ClassicMask}, called \textit{Bilateral Total-Variance (BTV)}, 

{\small
\begin{align}
        {\text BTV} = \sum_{u \in \Lambda} e^{-\nabla I(u)^2/\sigma^2} \| \nabla M(u)\|_{\beta}^{\beta} 
\end{align} \label{eq:btv}
}

\noindent where $M(u)$ and $I(u)$ are the mask and the input image value at pixel ($u$), and $\beta$ and $\sigma$ are hyperparameters. This enforces the mask to not only be smooth in its own space but also to consider the pixel value differences in the image space. This is intuitive since BTV would discourage mask value changes when the input image pixels have similar color. In other words, BTV penalizes the variation in the mask when it is over a single part of an object. This helps particularly in high-resolution mask optimizations and prevents having scattered and adversarial masks. 

% \vspace{-0.1in}
%----------------------------------
\subsubsection{Backtracking Line Search for Computing the Step Size }\label{sec:line_search}

Similar to \cite{IGOS}, we use backtracking line search at each iteration of mask update. 
%This 
%is due to the complexity of eq. (\ref{eq:igos++}) where fixed step sizes does not converge, especially there is no universal step size that converges under different regularization parameters. 
Appropriate step size plays a significant role in avoiding local optimum and accelerates convergence. To that end, we revised the Armijo-Goldstein condition as follows,

{\small
    \begin{align}
        & \sum_{s=1}^{S} F_c\bigg(\frac{s}{S} (M^k-\alpha^k \cdot TG(M^k))\bigg) - \sum_{s=1}^{S}F_c(\frac{s}{S}M^k) \notag \\
        \leq & -\alpha^k \cdot \beta \cdot TG(M^k)^{T}TG(M^k), 
    \end{align} \label{eq:revised_armijo}
}

\noindent where $\alpha^k$ is the step size at time step $k$ and $\beta \in (0, 1)$ is a control parameter. This attempts to determine the maximum movement in the search direction that corresponds to an adequate decrease in the objective function $F_c(M^k)$. Note, this is slightly different from the revised condition in \cite{IGOS} in that the objective function decrease is calculated over the IG intervals rather than at the mask $M^k$. This is due to the fact that IG actually solves an optimization problem similar to eq. (\ref{eq:ig_real}). 
\section{Experiments} \label{sec:exp}
% \vspace{-0.1in}
% -----------------------------------------------
In this section we will validate the algorithm, first through an experiment in the natural image domain in order to compare with other baselines. Then, we will show the application of the algorithm to the analysis of COVID-19 X-ray images.
\subsection{Metrics and Evaluation Setup}

% Despite the recent advances in the explainable AI research, there is no consensus about a universal metric to evaluate methods such as saliency maps in terms of \textit{interpretability}s. 
We first evaluate the algorithm in the natural image domain to validate it against other baselines. Although visual assessment of the saliency maps  might seem straight forward, quantitative comparisons still pose a challenge. For example, one of the widely-used metrics is the \textit{pointing game} \cite{zhang16excitationBP} which assesses how accurate the saliency maps can locate the objects using the bounding-boxes annotated by \textit{humans}. However, this is not necessarily correlated with the underlying decision-making process of the deep networks, and localization of the objects is only intelligible for humans. Also, pointing game requires correct localization even for \textit{misclassified} examples, and its results correlate poorly with other metrics (e.g. Excitation-BP, which has excellent pointing game scores, suffers heavily from the fallacies pointed out in \cite{adebayo2018sanity}). This casts doubt on the validity of this metric. Due to these flaws, we do not utilize this metric in our evaluations.
% Also, we have noticed that numerous samples, in both MSCOCO and PASCAL VOC datasets, used in pointing game experiments are \textit{misclassified} by the network. This casts doubt on the validity of this metric. Due to these flaws, we do not use this metric in our evaluations. 

We opted to follow the causal metrics introduced in \cite{2018RISE}, which is the evolved version of the "deletion game" introduced in \cite{ClassicMask}. Although not perfect, we find this to be a better evaluation metric, as it performs \textit{interventions} on the input image, a necessary approach to understand causality. These metrics are: the \textit{deletion metric} that evaluates how sharply the confidence of the network drops as regions are removed from the input. Starting from the original image, relevant pixels as indicated by the heatmap are gradually deleted from the image and are replaced by pixels from the highly-blurred baseline. This goes on until all pixels from the original image are removed and network has near-zero confidence. The deletion score is the area under the curve (AUC) of the classification confidences. The lower the deletion score is the better. The \textit{Insertion metric}, which is complementary to the deletion one, shows how quickly the original confidence of the network can be reached if relevant evidence are presented. Starting from a baseline image with near-zero confidence, relevant pixels from the image, based on the ranking provided by the heatmap, are gradually inserted into the baseline image. This goes on until all pixels in the baseline are replaced by the original image. The insertion score is also the AUC of the classification scores, with a higher insertion score indicating better performance. While adversarial examples can break the deletion metric with ease and achieve near perfect score, they usually do poorly on the insertion score. Yet only relying on the insertion score will sometimes invite irrelevant background regions to be ranked higher than relevant areas. That is why one needs to jointly look at both scores. See supplementary materials for the set of parameters used in our experiments.
%That is why solely relying on deletion metric might be misleading.

\begin{figure}[t!]
\begin{center}
{ 
\includegraphics[width=0.99\linewidth]{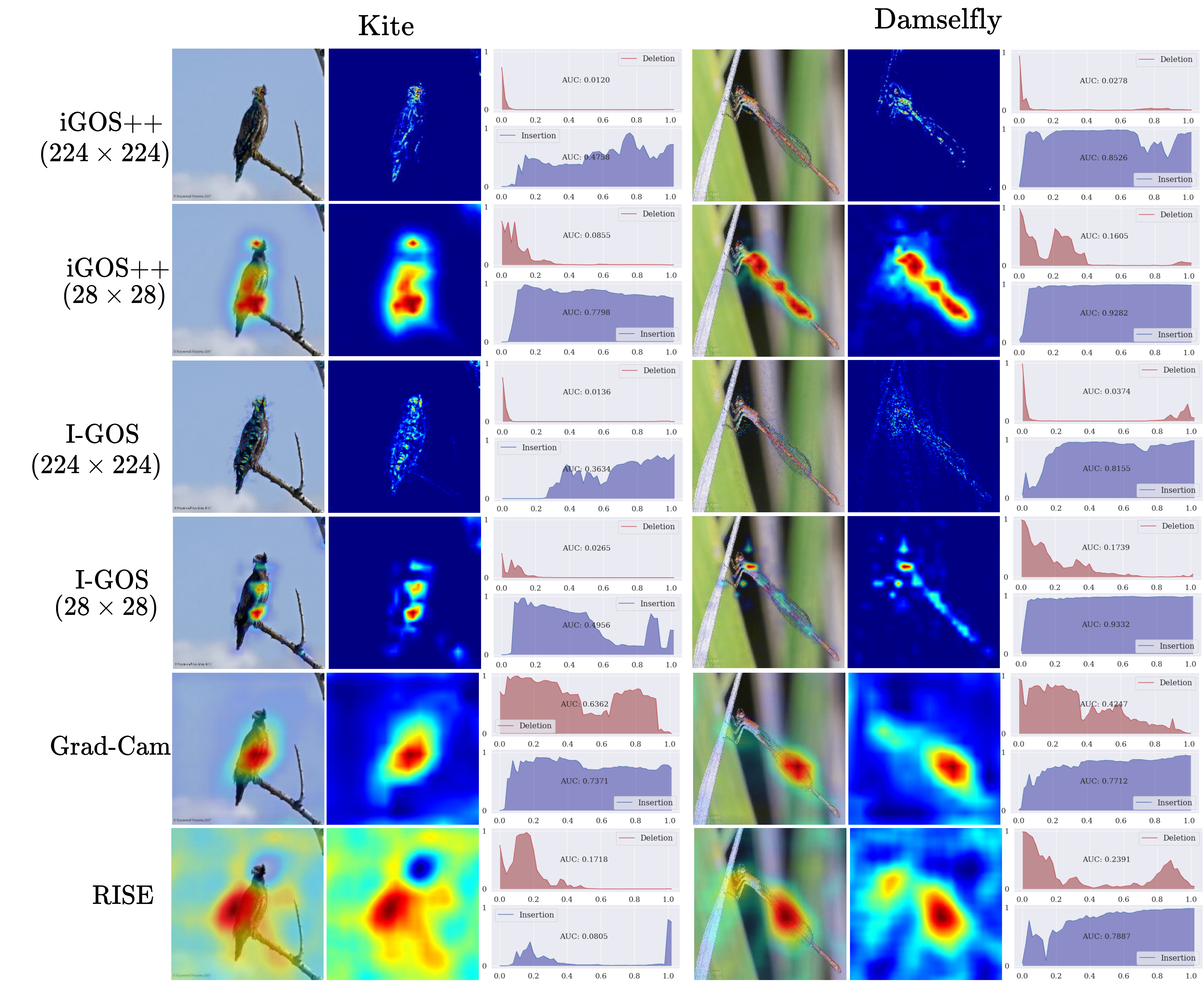}
}
\end{center}
% \vskip -0.15in
\caption{\small Heatmap visualization from different methods. It can be seen that iGOS++ has better insertion and deletion curves than the baselines. Note the difference in detail between the $28\times 28$ and $224 \times 224$ masks (Best viewed in Color) }\label{fig:vis}
%  \vskip -0.15in
\end{figure}

% \vspace{-0.1in}
%----------------------------------
\subsection{Results and Analysis}

%------------------------%------------------------
%  RESNET INSERTION DELETION
\begin{table*}[hbt!]
\caption{Quantitative comparison in terms of deletion (lower is better) and insertion (higher is better) metrics using ResNet50. The top row shows the different resolutions}
\label{table:resnet50}
% \vskip 0.05in
\begin{center}
\begin{small}
\begin{tabular}{l|cc|cc|cc}
\toprule
{\sc \bf ResNet50} & \multicolumn{2}{c|}{224$\times$224} & \multicolumn{2}{c|}{28$\times$28}  & \multicolumn{2}{c}{7$\times$7}\\
{} & Deletion & Insertion    & Deletion & Insertion       & Deletion & Insertion\\ 
\midrule

GradCam  \cite{Gradcam17}&  -- --& -- --      &  -- --& -- --    & 0.1675& 0.6521\\ 

Integrated Gradients \cite{IntegratedGradient} & 0.0907& 0.2921      &  -- --&  -- --     &  -- -- &  -- --\\ 

RISE \cite{2018RISE}& 0.1196 & 0.5637   & -- --& -- --     & -- --&  -- --\\ 

Mask \cite{ClassicMask}& 0.0468 & 0.4962         & 0.1151& 0.5559         & 0.2259 & 0.6003 \\ 

IGOS \cite{IGOS}& 0.0420 & 0.5846 &      0.1059 & 0.5986 &     \textbf{0.1607} & 0.6632 \\ 

% I-GOS + Insertion + BTV & 0.0372 & 0.7442 & 0.1041 & 0.7092 & 0.1870 & 0.7183 \\

iGOS++ (ours) & \textbf{0.0328} & \textbf{0.7261} &   \textbf{0.0929} & \textbf{0.7284} & \text{0.1810} & \textbf{0.7332} \\

\bottomrule
\end{tabular}
\end{small}
\end{center}
% \vskip -0.25in
\end{table*}

%------------------------%------------------------
%  VGG INSERTION DELETION
\begin{table*}[hbt]
\caption{Quantitative comparison in terms of deletion/insertion metrics using VGG19. The top row shows the different resolutions.}
\label{table:vgg19}
% \vskip 0.15in
\begin{center}
\begin{small}
\begin{tabular}{l|cc|cc|cc}
\toprule
{\sc \bf VGG19} & \multicolumn{2}{c|}{224$\times$224} & \multicolumn{2}{c|}{28$\times$28}  & \multicolumn{2}{c}{14$\times$14}\\
{} & Deletion & Insertion    & Deletion & Insertion       & Deletion & Insertion\\ 
\midrule

GradCam \cite{Gradcam17} & -- -- &  -- --       &  -- --&  -- --         & 0.1527 & 0.5938  \\ 

Integrated Gradients \cite{IntegratedGradient} & 0.0663 & 0.2551          &  -- -- &  -- --       &  -- -- &  -- -- \\ 

RISE  \cite{2018RISE} & 0.1082 & 0.5139       &  -- -- &  -- --      &  -- --&  -- --  \\ 

Mask \cite{ClassicMask}& 0.0482& 0.4158        & 0.1056 & 0.5335          & 0.1753 & 0.5647  \\ 

I-GOS \cite{IGOS}& \textbf{0.0336} & 0.5246     & 0.0899 & 0.5701       & \text{0.1213} & 0.6387 \\ 
% I-GOS + Insertion + BTV & 0.0356 & 0.6412 & 0.0896 & 0.6385 & 0.1725 & 0.6774 \\

iGOS++ (ours) & \text{0.0344} & \textbf{0.6537}       & \textbf{0.0796} & \textbf{0.7066}       & \textbf{0.1123} & \textbf{0.7061} \\

\bottomrule
\end{tabular}
\end{small}
\end{center}
% \vskip -0.25in
\end{table*}

%---------------------------------------------------------

\textbf{Insertion and Deletion Scores.}
Table \ref{table:resnet50} and Table \ref{table:vgg19} show our results on the Insertion and Deletion metrics, averaged over $5,000$ random ImageNet images, for ResNet50 and VGG19 architectures respectively. We experimented with masks of the size  $224 \times 224$, $28 \times 28$, $7 \times 7$ (only on Resnet50), and $14 \times 14$ (only on VGG19) in our experiments. The choice of the masks is to provide fair comparison with other baselines, e.g GradCam \cite{Gradcam17} that for ResNet50 is best on its``\textit{layer4}'' ($7 \times 7$ resolution). In addition, Integrated Gradients visualization \cite{IntegratedGradient} has heatmaps directly on the input ($224 \times 224$). Also note that binary masks for RISE \cite{2018RISE} are generated at $7 \times 7$ and then up-sampled to $224 \times 224$.

As shown in Tables \ref{table:resnet50} and \ref{table:vgg19}, iGOS++ has superior performance at all resolutions compared to the other approaches. Particularly, for the insertion score, our approach is showing $10\%-25\%$ improvement over prior work. This is mainly due to the novel  incorporation of the insertion objective in our eq. (\ref{eq:igos++}). Note that, for all the methods, the insertion score tends to be lower on higher resolution heatmaps, since it is easier for CNNs to extract features if a significant contiguous part is inserted, rather than isolated pixels in high-resolution masks. On the other hand, on higher resolutions, it is easier to break an image feature by destroying a small number of pixels, hence the deletion metric is usually better. We note that the drop of our insertion performance from $7\times 7$ (or $14\times 14$) to $224 \times 224$ is significantly smaller than that of I-GOS, hence it is safer to choose a higher-resolution with iGOS++. Note, we did not notice high variance in our reported results. For instance, on the reported numbers in Table \ref{table:resnet50} for iGOS++ ($224 \times 224$ resolution), we obtained the standard deviations of $0.032 \pm 0.00023$ and $0.722 \pm 0.00097$) for the insertion and deletion scores, respectively. The low variance shows validity of the experiment setup.

Furthermore, Fig.\ref{fig:vis} shows visualizations from multiple saliency map methods along with their corresponding deletion and insertion curves and scores. Saliency map for I-GOS and iGOS++ are generated at high ($224\times224$) and medium ($28\times28$) resolutions. It can be seen that iGOS++ performs well even if the object has long and thin parts, whereas I-GOS generates much more scattered heatmaps at high resolutions. GradCAM works well on insertion but included too many irrelevant regions due to its low resolution, a similar issue shared with RISE. For additional visual comparison of our method against various baselines please refer to the supplementary materials.

% \textbf{Visual comparison}

%$28\times 28$ is very small, hence one can always safely choose at least a $28 \times 28$ resolution, if not even higher, whereas in I-GOS the drop in insertion metric from $7\times 7$ to $28\times 28$ is significant.

%It is not proper to choose from the resolutions based on their scores. The choice for the ``best'' mask depends on the application and it offers a trade off between the smoothness and the amount of details one requires in the saliency map. Note, typically the lower the resolution, the smoother the mask is. 

%------------------------
%  SANITY CHECK FIGURE
\begin{figure}[h!]
\begin{center}
{ 
\includegraphics[width=0.99\linewidth]{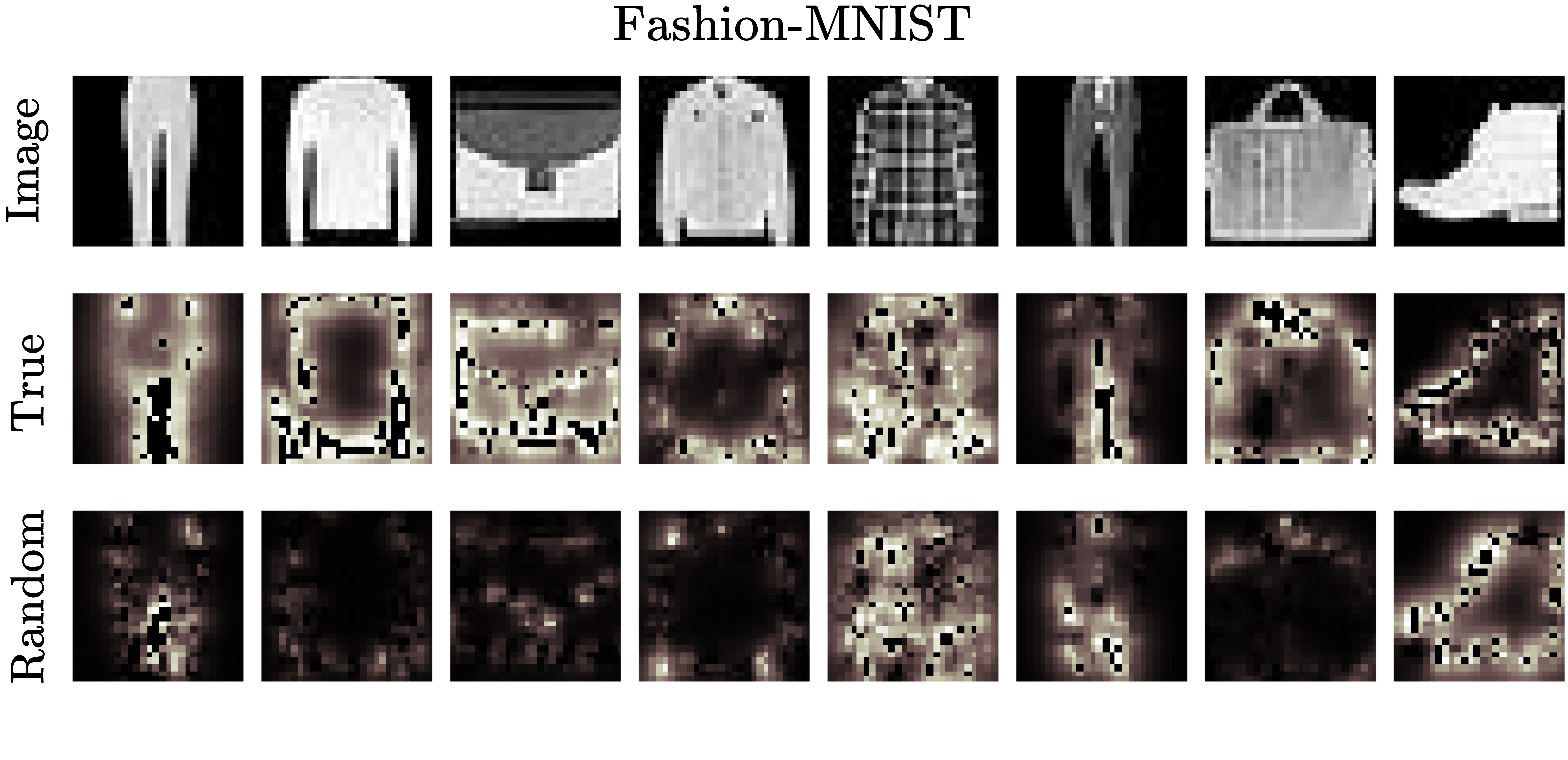}
}
\end{center}
%  \vskip -0.15in
 \caption{Sanity check using the Fashion MNIST dataset. A simple CNN model was trained first using ground truth labels and then with random labels. The masks on the second row were generated with the ground truth trained model, while the third row are masks generated for the random label trained model }
 \label{fig:sanity09}
\end{figure}

%------------------------
\begin{figure}[h!]
\begin{center}
\subfloat[]{\includegraphics[width=0.99\linewidth]{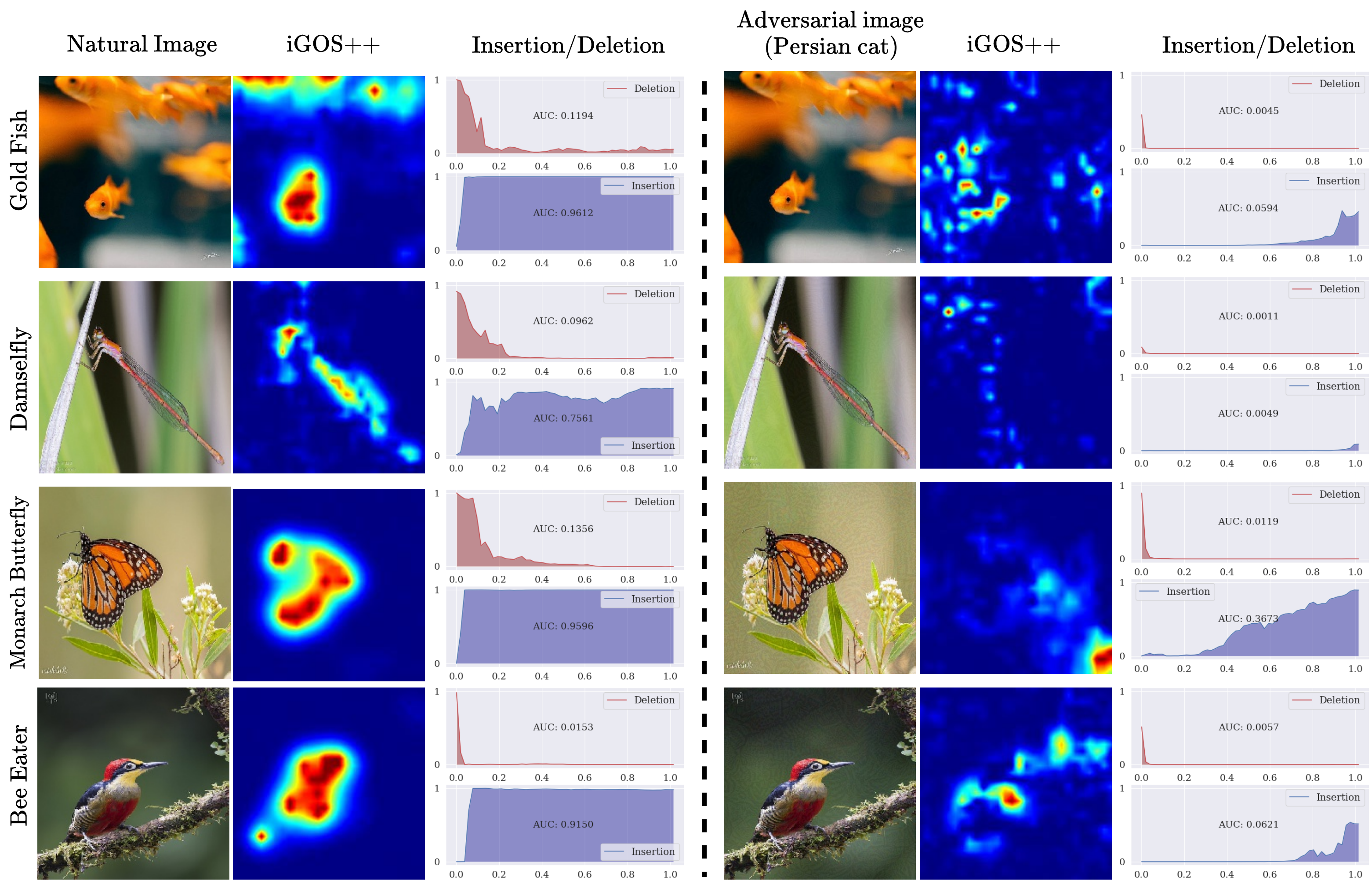}
}
\end{center}
\begin{center}
    \subfloat[]{\includegraphics[width=0.99\linewidth]{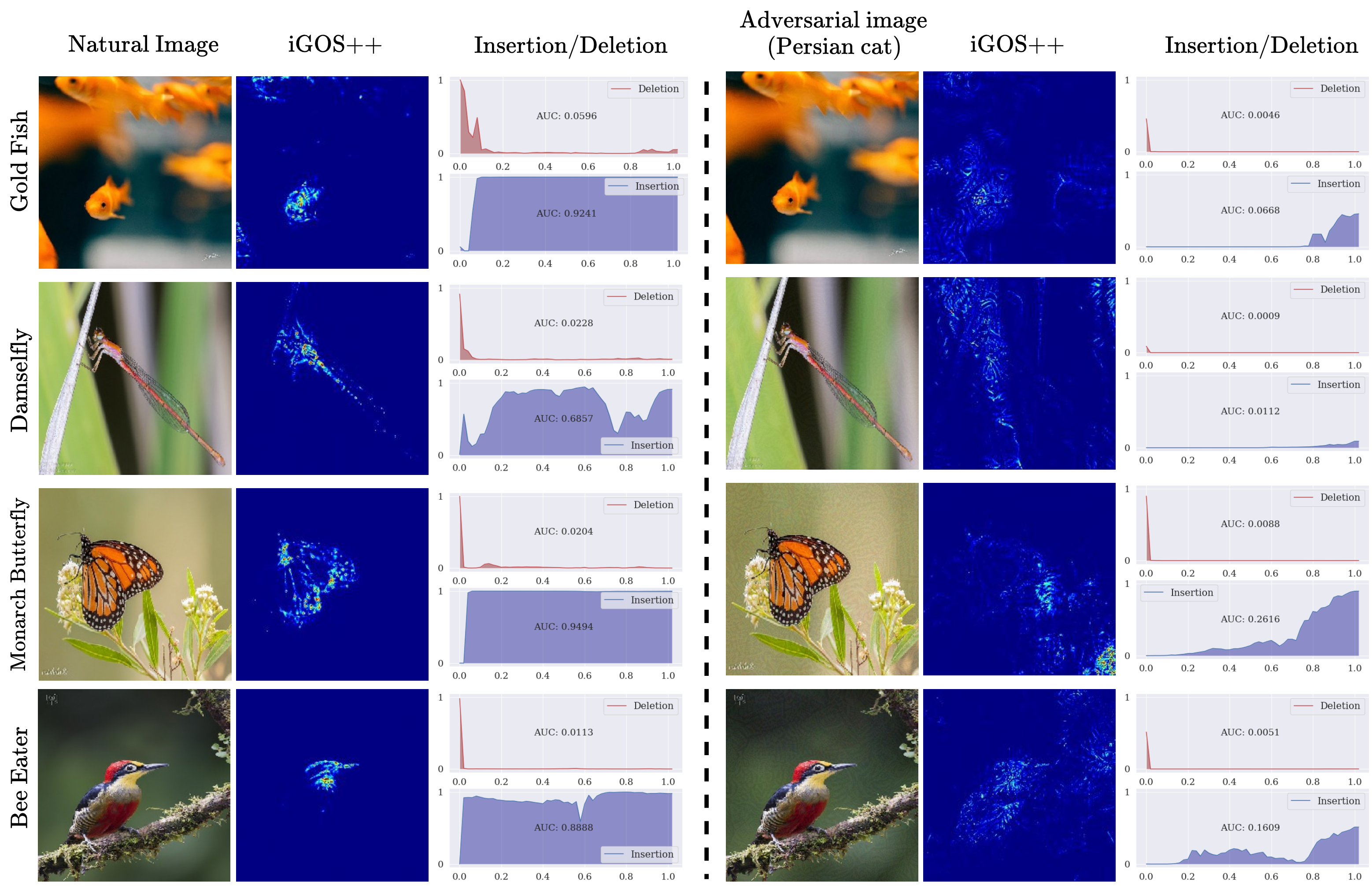}
}
\end{center}
\caption{Comparison of heatmap visualizations from iGOS++ for natural and adversarial examples on two different resolutions: a) 28$\times$28 b) 224$\times$224. The insertion curve only goes up at the end on adversarial examples, indicating that the full image is needed to explain adversarial predictions, while for natural images, the insertion curve goes up quickly. The heatmaps are also different, and for adversarial examples, they do not represent \textit{meaningful} explanations. }
% \vskip -0.05in 
\label{fig:adv}
\end{figure}

%---------------------------------------------------------
\textbf{Sanity Check.}
Following the work of \cite{adebayo2018sanity}, we perform sanity checks on our method to justify its validity, as only relying on visual assessment can be fallacious. First is the \textit{model randomization} test which checks whether the generated heatmaps are indeed independent of the model parameters or not. Since by randomizing the model (even over one layer), the output score of the network will go near-zero, our method simply returns the initialization $M\approx{\bf 1}$ which is different from the final mask we generate. Second, for the \textit{label randomization} test, we trained a simple Convolutional Neural Network (CNN) over the Fashion-MNIST dataset.
The CNN was trained with two sets of data.
First it was trained using the ground truth labels of the dataset until it reached 100\% accuracy.
Heatmaps were generated for this model and then compared to maps generated for a model trained with random labels.
The idea is if the heatmaps are the same, then the method might be interpreting the image rather than explaining the models decision.
Figure \ref{fig:sanity09} shows that iGOS++ generates meaningful heatmaps on real classes that are significantly different from random classes.

%---------------------------------------------------------
\textbf{Adversarial Examples}
In addition to the sanity check presented in \cite{adebayo2018sanity}, the visualization from our method on adversarial examples have been analyzed. Figure \ref{fig:adv} shows four image samples from ImageNet Validation set where the heatmaps generated on the adversarial images have been visualized along with the original natural images. In this experiment, VGG19 architecture and two different mask resolutions: a \emph{low-resolution} ($28 \times 28$) and a \emph{high-resolution} ($224 \times 224$) are used. To generate adversarial examples (target class: \textit{persian cat}), we used the MI-FGSM \cite{Dong_2018_CVPR} on the VGG19 model. There are two main takeaways from this analysis: \emph{first}, for adversarial examples, the heatmaps generated by iGOS++ does not provide \textit{meaningful} explanations. \emph{Second}, it can be observed that for adversarial examples, the insertion score is significantly lower and it only reaches the original score after almost all the pixels have been inserted. This shows that the insertion score is a good indication of whether the generated mask is adversarial or not.

%---------------------------------------------------------

%  ABLATION 
\begin{table}[hbt!]
\caption{\small Results from ablation study on ResNet50. The top row shows the different mask resolutions.}
\label{table:ablation}
% \vskip 0.15in
\begin{center}
\begin{small}
\begin{tabular}{l|cc|cc}
\toprule
{\sc \bf Ablation} & \multicolumn{2}{c|}{224$\times$224} & \multicolumn{2}{c}{28$\times$28} \\
{} & Deletion & Insertion    & Deletion & Insertion \\ \hline

I-GOS                          &  0.0420   &     0.5846       & 0.1059 &  0.5986                     \\ \hline
Insertion                      &  0.0760   &     0.6192       & 0.1321 & 0.7231                         \\
I-GOS + Insertion (na\"{i}ve)  &  0.0322   &     0.6175       & 0.2037 & 0.5103                                           \\
% I-GOS + Insertion + BTV  &  0.0372   &     0.7442       & 0.1041 & 0.7092                                           \\
iGOS++ (no noise)              &  0.0490   &     0.5943       & \text{ 0.0904} & 0.7108                         \\
iGOS++ (fix step size)         &   0.0332  &     0.5695       & 0.1052 & 0.7060                     \\
iGOS++ (no BTV)                   &  \textbf{0.0245}       &    0.6742    &  \textbf{0.0813} &  0.6825                                                             \\ \hline
iGOS++                        &  0.0328 & \textbf{0.7261}    & 0.0929 &  \textbf{0.7284}        \\

\bottomrule
\end{tabular}
\end{small}
\end{center}
% \vskip -0.15in
\end{table}

\textbf{Ablation Study.} 
The results from the ablation study are presented in Table \ref{table:ablation}. The experiments are run at $224 \times 224$ and $28 \times 28$ resolutions using the ResNet50 model on ImageNet. We observe that optimizing a mask by replacing the deletion loss in \cite{IGOS} with an insertion one obtains a good insertion score on both resolutions. However, it significantly hurts the deletion score. On the other hand, na\"{i}vely incorporating the insertion loss to eq. (\ref{eq:igos}) by just adding an insertion loss term clearly does not work as well. In fact, it is performing worst than either of the \cite{IGOS} and insertion optimization alone. Further, we find that adding noise makes a clear difference in high resolution ($224 \times 224$). Moreover, having a fixed step size is worse than using an adaptive step size with line-search. Finally, by removing the bilateral TV term we observed that the deletion score decreases, particularly in high resolution while the insertion score also was impacted negatively. This shows the benefit of the bilateral TV term in avoiding adversarial solutions. In addition, for a comparison between the deletion and insertion masks, please refer to the supplementary materials. %The addition of the bilateral TV term also helps significantly the simple I-GOS + insertion baseline however, it still performs slightly worse than iGOS++ in most cases.

%---------------------------------------------------------
\textbf{Convergence Behavior.}
Throughout our experiments, we found the convergence behavior of our method robust against the choice of hyperparameters. The choice of hyperparameters were also transferable between all the datasets and models as stated in the \textit{Set of Hyperparameters} section. Table. \ref{table:objective}, shows the final value of the combined deletion and insertion loss $f_c\big(\Phi(I_0,\tilde{I}_0, M)\big) - f_c\big(\Phi(I_0,\tilde{I}_0, 1 - M)\big)$ (eq. (4) in the main paper) after optimizing with iGOS++ and the na\"{i}ve extension of I-GOS \cite{IGOS} --- when the insertion loss is directly added to it. The reported numbers are averaged over 500 images for different choices of hyperparameters $\lambda_1$ and $\lambda_2$. Although, the na\"{i}ve extension is directly optimized over the combined deletion and insertion loss, iGOS++ shows superior performance in all settings. Note the loss value can be negative since the insertion loss is maximized during the optimization. This validates the capability of iGOS++ in achieving a lower objective compared to using the naive extension of I-GOS for the same objective, showing that it found better optima in our difficult non-convex optimization problem. Running time comparison of iGOS++ is provided in the supplementary materials.

\begin{table}[htb!]
% \vskip -0.15in
\caption{Optimization objective comparison with naive addition of the insertion loss at 28$\times$28 resolution, averaged over 500 images. Our algorithm resulted a lower objective than the naive version for the same objective, showing that it found better optima in this difficult non-convex optimization problem. Note that $\lambda_2 = 200$ was never used in the actual experiments and only included here for completeness. Also note the loss could be negative because there is a negative sign on the insertion loss in eq. (4) in the main paper}

\label{table:objective}
% \vskip -0.05in
\begin{center}
\begin{small}
\begin{tabular}{c|c|c|c|c}
\toprule
{\bf Loss Value} & {$\lambda_1$=1} & {$\lambda_1$=1} & {$\lambda_1$=10} & {$\lambda_1$=10} \\
{\bf} & {$\lambda_2$=2} & {$\lambda_2$=20} & {$\lambda_2$=20} & {$\lambda_2$=200} \\
\hline 
I-GOS + Ins (na\"{i}ve)  & {-0.869} & {-0.374}  & {0.346} & {0.693} \\
iGOS++ & \textbf{-0.971}  &  \textbf{-0.649} & \textbf{0.158} & \textbf{0.654} \\
\bottomrule
\end{tabular}
\end{small}
\end{center}
% \vskip -0.15in
\end{table}

\subsection{COVID-19 Detection from X-ray Imaging} 
COVID-19 has been devastating to human lives throughout the world in 2020. Currently, diagnostic tools such as RT-PCR has non-negligible false-negative rates hence it is desirable to be able to diagnose COVID-19 cases directly from chest imaging. X-ray imaging is significantly cheaper than CT or other more high-resolution imaging tools, hence there would be significant socio-economical benefits if one can diagnose COVID-19 reliably from X-ray imaging data, especially in an explainable manner. %Especially, we would hope that by explaining automatic deep classifiers that enjoy a high accuracy, we could teach humans a thing or two about the diagnosis of this new disease. 
To that end, we used the COVIDx dataset \cite{wang2020covidnet} which is one of the largest publicly-available COVID-19 dataset with 13,786 training samples and 1,572 validation samples comprised of X-ray images from \emph{Normal}, \emph{Pneumonia}, and \emph{COVID-19} patients. Following the setting from \cite{wang2020covidnet}, we trained a classifier over these images.

\begin{figure}[hbt!]
\begin{center}
{ 
\includegraphics[width=0.99\linewidth]{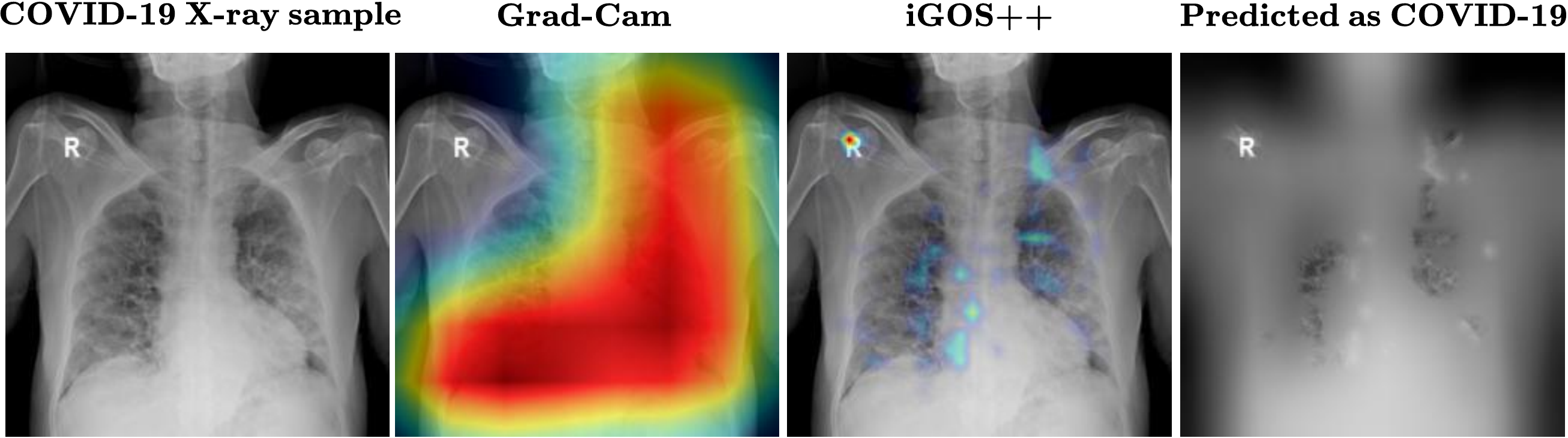}
}
\end{center}
%  \vskip -0.05in
 \caption{\small Showcase of the capability of iGOS++ in detecting bugs in a COVID-19 classification pipeline. It can be noted that unlike Grad-Cam which provides coarse (8$\times$8 resolution) and ineffective explanation, iGOS++ can generate a more-detailed explanation (32$\times$32 resolution) to discover the most salient regions. In the right-most image, only by inserting the top 6\% of the pixels from iGOS++ heatmap (highlighting the character ``R"), the classifier predicts it as COVID-19 (confidence 43 \%). }
 \label{fig:covid}
\end{figure}

Somewhat surprisingly, when we applied iGOS++ on the trained classifier, we noticed that in occasional cases the classifier seems to have overfitted to singleton characters printed on the x-ray image (Fig.~\ref{fig:covid}), such that even when only the character region is available, there is a non-negligible chance of classifying for COVID-19. Note that the higher resolution explanation from iGOS++ is important in pinpointing the heatmap to the character whereas low-resolution alternatives such as GradCAM were not informative. For further examples and the case when only text region is revealed to the classifier, please refer to the supplementary materials. 

%------------------------

Noticing this ``bug" of the classifier, we utilized a state-of-the-art character detector, CRAFT \cite{baek2019character}, and removed the spurious characters from all the x-ray images in both the training and the testing sets by cleaning and in-painting the detected regions. Examples from the original and cleaned dataset, referred to as \textit{COVIDx++}, can be found in the supplementary materials. The results shown in Table~\ref{table:covid} show that the recall of COVID-19 detection on the validation dataset improved by $2.5\%$ and the F1 score improved by more than $1\%$. This small exercise showcases that ``bugs" do exist in deep network classifiers as they do not have common sense on what part of the data is definitely noise, and that heatmap visualizations can help humans locate these bugs as a useful debugging tool. We hope to dig more into this experiment in the future and obtain more meaningful knowledge from this data.

%------------------------
\begin{table}[htb]
\caption{\small Classification performance on the validation set of the COVIDx and COVIDx++ datasets. COVID-NET \cite{wang2020covidnet} is used as the classifier.}
\label{table:covid}
% \vskip 0.15in
\begin{center}
\begin{small}
\begin{tabular}{c|c|c|c|c}
\toprule
{Dataset } & {Accuracy} & {F1-Score} & {Precision} & {Recall} \\
\hline
{COVIDx} & {95.19} & {93.81} & \textbf{95.75} & {91.85} \\
{COVIDx++} & \textbf{95.93} & \textbf{95.08} & {95.70} & \textbf{94.49}\\
    
\bottomrule
\end{tabular}
\end{small}
\end{center}
%\vskip -0.15in
\end{table}

% %------------------------
% \begin{figure}[hbt!]
% \begin{center}
% { 
% \includegraphics[width=0.95\linewidth]{imgs/inpaint.jpg}
% }
% \end{center}
% % \vskip -0.15in
%  \caption{Example of X-ray images from COVID-19 patients in the COVIDx dataset (left column), the detected bounding boxed using the CRAFT text detector pipeline (middle column), and the final inpainted images (COVIDx++).}
%  \label{fig:inpaint}
% \end{figure}

% %-------------------------
% \textbf{D. \quad}\textbf{Data Cleaning for the Generation of The COVIDx++ dataset}: The first stage in cleansing the COVIDx dataset is to detect the text in the X-ray images. For this purpose, the CRAFT text detection method \cite{baek2019character} is used. We used the general pre-trained model available at the official code repository \footnote{\url{https://github.com/clovaai/CRAFT-pytorch}}. The text confidence threshold of 0.7 is used for the experiments. To inpaint inside the bounding boxes detected from the CRAFT, we used the built-in OpenCV inpainting function ({\texttt{cv2.inpaint(...)}}) with the algorithm by \cite{telea2004image} and inpaint radius of 10 pixels. Figure \ref{fig:inpaint} shows some samples from the COVID-19 patients in  the COVIDx dataset \cite{wang2020covidnet}, the corresponding detected bounding boxes detected by the CRAFT text detector, and the final inpainted images. \\

\begin{figure}[hbt]
\begin{center}
{
\includegraphics[width=0.99\linewidth]{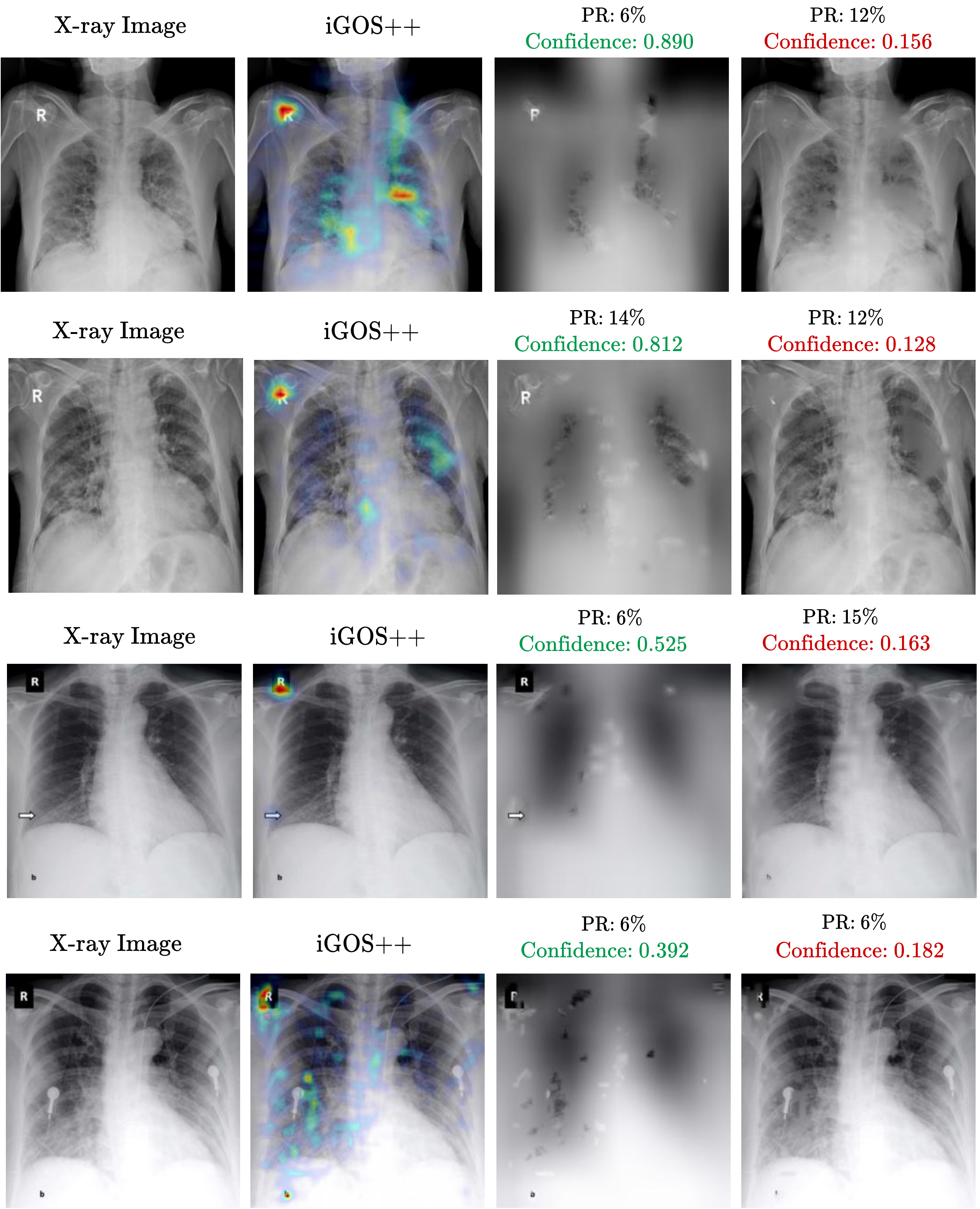} 
}
\end{center}
% \vspace{-.15in}
\caption{Showcase for capability of the iGOS++ in detecting bugs in a COVID-19 classification pipeline. The X-ray images in the first column belong to \text{COVID-19} patients. The second column shows iGOS++ visualizations at 32$\times$32 resolution where our method highlights the text region, the letter "R". The third column shows images when only a small fraction of the pixels (e.g. 6\%), referred to as Pixel Ratio (PR), are inserted back into the baseline. Interestingly, the insertion of these small amount of pixels, which mostly include the text region, is enough for the perturbed images to be classified as \text{COVID-19} (The corresponding confidence scores are shown in color green). The fourth column shows the opposite scenario where only a small fraction of the pixels (e.g. PR: 12\%) are removed (blurred) from the original image and that is enough for the perturbed images to be \textit{mis-classified}, i.e. not \text{COVID-19}. Obviously, the confidence score for these images (shown in color red) are lower than of \text{Normal} or \text{Pneumonia} classes.}
\vspace{-.15in}

\label{fig:covid_supp}
\end{figure}

Figure. \ref{fig:covid_supp} showcases further examples of our visualization method giving insights to the underlying decision making process of the classifier and potentially debuging and improving it. It can be seen that the iGOS++ visualizations (second column) are highly weighting in the text region in the images, i.e. the letter "R". The third column shows images when only a small fraction of the pixels (e.g. 6\%), referred to as \textit{Pixel Ratio} (PR), were inserted back into the baseline image (a Gaussian-blurred of the original image). However, these small amount of pixels, which mostly include the text region, are predicted as \text{COVID-19} classes with their corresponding confidence (shown in green color). On the other hand, using the same iGOS++ visualizations, when only a small fraction of the pixels (e.g. PR: 12\%) are removed from the original image, the classifier shows different predictions (than \text{COVID-19}). The corresponding confidence (shown in red color) is depicted on the top of the perturbed images. It should be noted that this confidence is not high enough so that the images get classify as \text{COVID-19}.

\begin{figure}[ht!]
\begin{center}
\subfloat[]{\includegraphics[width=0.49\linewidth]{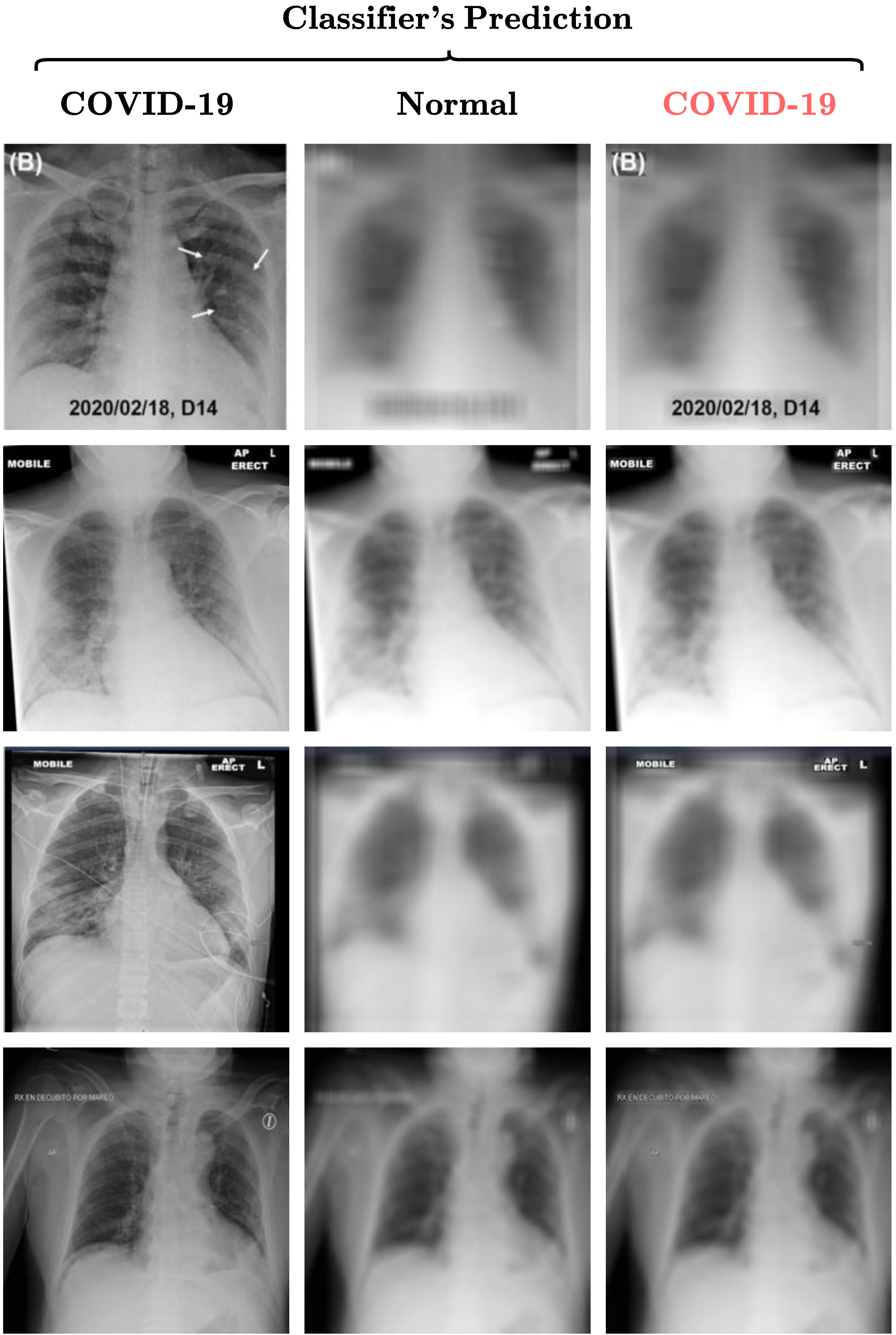}\label{fig:cvd_as_cvd}
}
\subfloat[]{\includegraphics[width=0.49\linewidth]{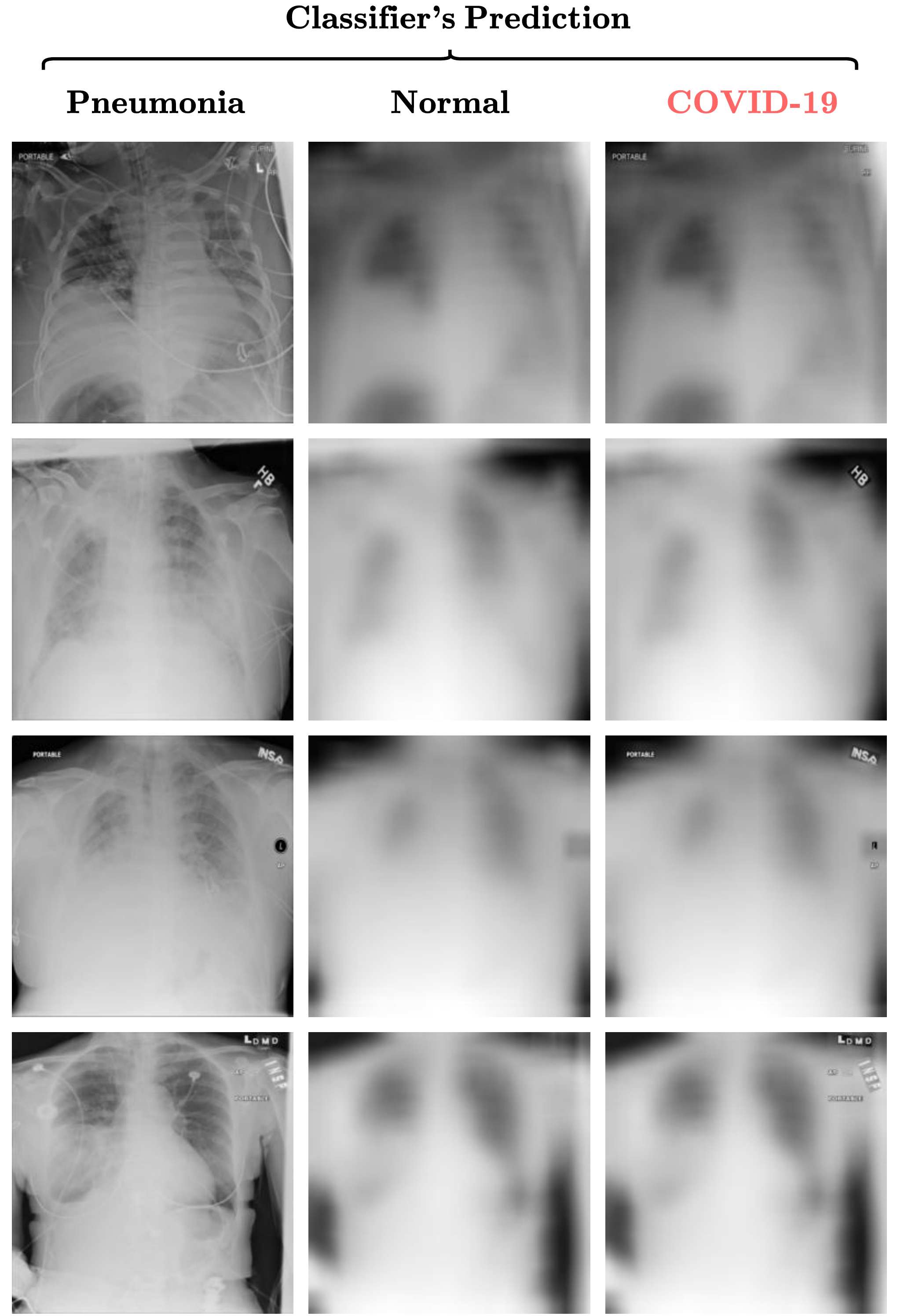}\label{fig:pne_as_cvd}
}
\end{center}
% \vspace{-.15in}
\caption{\small
The prediction of the classifier for each image is shown on the top. X-ray images of a) \textit{COVID-19} b) \textit{Pneumonia} patients from the COVIDx dataset \cite{wang2020covidnet} are depicted in the left columns. When the highly-blurred version of these images are fed to the classifier, they are all predicted as \textit{Normal} (middle columns). However, when only the text regions were inserted back into the baselines, the classifier falsely predicts all those images as \text{COVID-19}, shown by red color (third columns).This suggests that the text regions are directly influencing the decision making of the classifier.}
\label{fig:covid_text}
\end{figure}

Figure \ref{fig:covid_text} shows examples of the case when the sole insertion of the \textit{text region} into a baseline is enough for the classifier to change its prediction, \textit{falsely to \text{COVID-19}}. For the purpose of this analysis, we used a few examples of X-ray images from \text{COVID-19} (Fig. \ref{fig:cvd_as_cvd}) and \textit{Pneumonia} (Fig. \ref{fig:pne_as_cvd}) patients, depicted in the left columns. When feeding the highly-blurred version of these images to the classifier, they were all predicted as \textit{Normal} (the second columns). However, when only revealing the text regions back into the baselines (using the bounding boxed from the CRAFT text detector \cite{baek2019character}), the classifier falsely predicts all those images as \text{COVID-19}, shown by red color (third columns). This suggests that the presence of text regions is influencing the decisions of the classifier --- a bug that we hypothesized using our visualizing method. Nevertheless, in reality, one assumes these should be regarded as noise in the data and should be discarded by the classifier. This is an example of the fragility of trusting the decision making of \textit{black-box} classifiers, particularly in tasks where human life is at stake such as medical diagnosis.

\section{Conclusion and Discussion} \label{sec:conclusion}

We propose a new approach, iGOS++, for creating a heatmap visualization to explain decisions made by the CNNs. % that improves upon previous methods.
%Our approach uses integrated gradients and optimizes for saliency map.
%It uses a line search method to take full advantage of the properties of integrated gradients.
iGOS++ differs from other approaches by optimizing for separate deletion and insertion masks which are tied together to output a single explanation. We show empirically that with this approach, significantly better insertion performance at all resolutions can be achieved. % retain more information that is important to the network's decision.
Besides, we introduce a new term for regularization using the bilateral total variance of the image. This is shown to improve the smoothness of the generated heatmaps as well. As a real-life example, we showed that in a task of classifying COVID-19 patients from x-ray images, sometimes the classifier would overfit to the characters printed on the image. Removing this bug improved the classifier performance meaningfully. 
We hope the high-fidelity heatmaps generated with iGOS++ will be helpful for downstream tasks in explainable deep learning in both the natural image domain and medical imaging domain.
%We perform experiments to measure the deletion and insertion scores using two CNN models.

\section*{Acknowledgements}
We thank Dr. Alan Fern for helpful discussions. This work is supported in part by DARPA contract N66001-17-2-4030.

%-------------------------------------------------------------------------
\bibliographystyle{style}
\bibliography{main}

\clearpage
%-------------------------------------------------------------------------
% Document starts here
\section*{Supplementary Materials}

%---------------------------------------------------------
\textbf{A. \quad}\textbf{Deletion ($\bf M_x$) v.s. Insertion ($\bf M_y$) Mask.} $M_x$ and $M_y$ masks are optimized with different objectives and exhibit differences. To make the behavior of each of the $M_x$ and $M_y$ masks more clear, the insertion and deletion scores for each of them are reported in Table \ref{table:m_xy}. As backed by the results, $M_y$ (insertion mask) tends to be more local and smooth while $M_x$ is often more scattered. Multiplying them together, as in iGOS++ methodology, reduces adversariality and directs the optimization out of saddle points that come with the combination of the dueling loss functions as mentioned previously. In addition, it can be observed that in lower resolution (28$\times$28), $M_x$ is less adversarial than in higher resolution (224$\times$224). 
% These differences can also be observed in Fig. \ref{fig:m_xy} where the deletion mask $M_x$ is more adversarial and scattered while the insertion mask $M_y$ tends to highlight the "whole" (part of the) object. 

%------------------------
\begin{table}[htb]
\caption{Comparison of the Insertion/Deletion scores from iGOS++ with $M_x$ and $M_y$ masks at two different resolutions using ResNet50.}
\label{table:m_xy}
% \vskip -0.15in
\begin{center}
\begin{small}
\begin{tabular}{l|cc|cc}
\toprule
{\sc \bf $\bf M_x$ \& $\bf M_y$} & \multicolumn{2}{c|}{224$\times$224} & \multicolumn{2}{c}{28$\times$28} \\
{} & Deletion & Insertion    & Deletion & Insertion \\ \hline
$M_x$                   &  \textbf{0.0268}       &    0.5008          &  {0.1011}  &  0.5536  \\
$M_y$                   &  {0.0594}              &    0.7184          &  {0.1788}  &  0.6912  \\
$M_{xy}$(iGOS++)         &  0.0328                & \textbf{0.7261}    & \textbf{0.0929}            &  \textbf{0.7332}    \\
 
\bottomrule
\end{tabular}
\end{small}
\end{center}
%\vskip -0.15in
\end{table}

%------------------------
% \begin{figure}[hbt]
% \begin{center}
% { 
% \includegraphics[width=0.95\linewidth]{imgs/m_xy.pdf}
% }
% \end{center}
% % \vskip -0.15in
%  \caption{Visual comparison of $M_x$ and $M_y$ against iGOS++. $M_x$ is more scattered and adversarial while $M_y$ is showing coherent "whole" (part of the) object but encompassing less details.}
%  \label{fig:m_xy}
% \end{figure}

%---------------------------------------------------------
\textbf{B. \quad}\textbf{Failure Case}
In Fig.1, we show an example of where our method does not work properly. This image is incorrectly predicted by the network and has low prediction confidence. As it can be observed, an adversarial mask has been generated from our method as the insertion curve requires almost all pixels to be inserted to get to the original confidence. Generally, when the initial prediction confidence is low, or the network is predicted incorrectly, our method does not work very well. %However, when the initial prediction confidence is low, one potential choice to mitigate this issue is to optimize the masks over the \textit{logits} rather than on the confidence scores after the softmax layer.  \\

%------------------------
\begin{figure}[hbt]
\begin{center}
{ \label{fig:fail}
\includegraphics[width=0.99\linewidth]{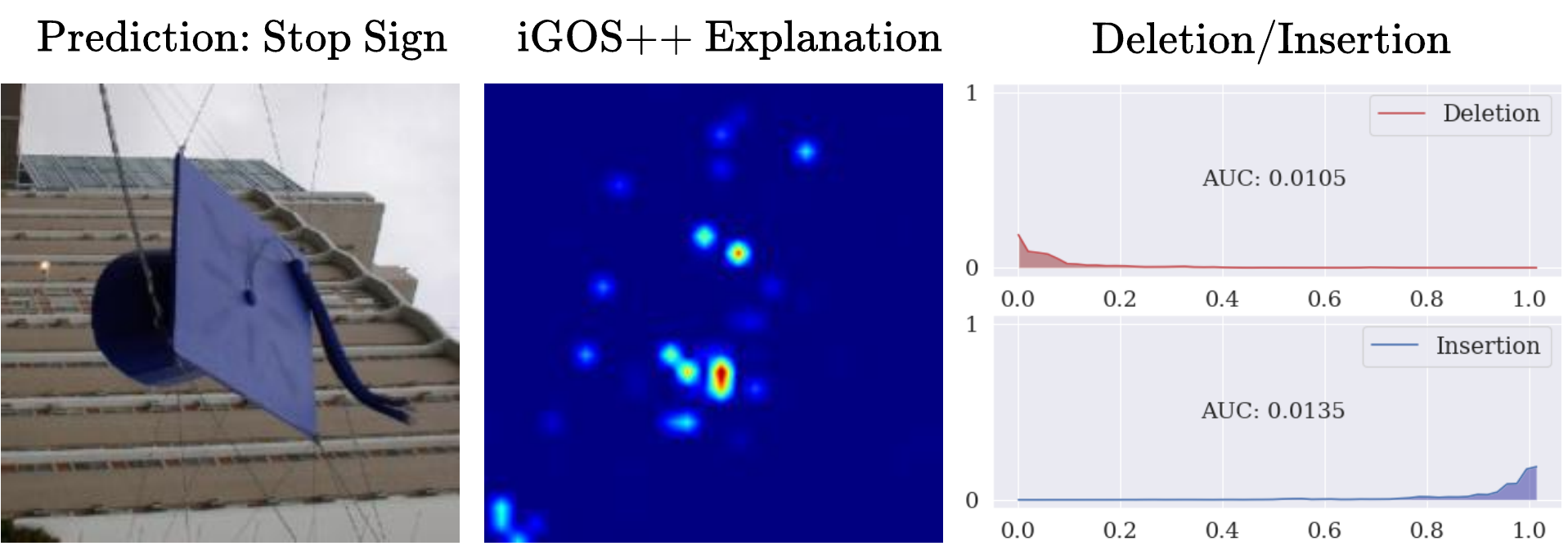}
}
\end{center}
\caption{\small Failure case for our method by finding and adversarial mask on $28\times28$. When the network initially has low prediction confidence, our method does not work well. It can be observed that the insertion score does not go up until near the end.}
 \vskip -0.1in
\end{figure}

%---------------------------------------------------------
\textbf{C. \quad}\textbf{Set of Hyperparameters}
(Insertion/Deletion): To quantitatively evaluate our method (iGOS++) against baselines in terms of the casual metrics, insertion and deletion scores, we use ImageNet \cite{ImageNet} benchmark. For the reported deletion/insertion scores, we use ResNet50 \cite{ResNet} and VGG19 \cite{VGG19} pre-trained on the ImageNet (from the PyTorch model zoo \cite{paszke2017automatic}) and generate heatmaps for 5,000 randomly selected images from the ImageNet validation set. All the baseline results presented in this paper are either obtained from the published paper when applicable or by running the publicly available implementations. The choice of hyperparameters is also chosen to have the best performance --- from the paper/code repositories. During all experiments, for high-resolution mask (224$\times$224), we set $\lambda_1=10$, and set it to 1 for all other resolutions. This is to avoid having a diffuse mask as the penalty over the mask size can easily be larger at high resolutions. In addition, $\lambda_2$ is set to $20$ for all resolutions. The line search parameters are similar to \cite{IGOS}. Also, We set $\beta=2$ and $\sigma=0.01$ for the BTV term. BTV and TV term (with $\beta=2$) also can be averaged and added to the main objective. Note that throughout all the experiments presented in the paper, the \textbf{same} set of hyperparameters is used for \textit{all network architectures} (ResNet50, VGG19, COVID-NET, etc.) as well as \textit{all datasets} (ImageNet, Fashion-MNIST, COVIDx, etc.), showcasing the robustness of the algorithm. We used the publicly-available implementation for COVID-NET  \footnote{\url{https://github.com/velebit-ai/COVID-Next-Pytorch}} and the best F-1 score from the validation set is reported in the paper.\\

%------------------------
\begin{figure*}[hbt!]
\begin{center}
{ 
\includegraphics[width=0.95\linewidth]{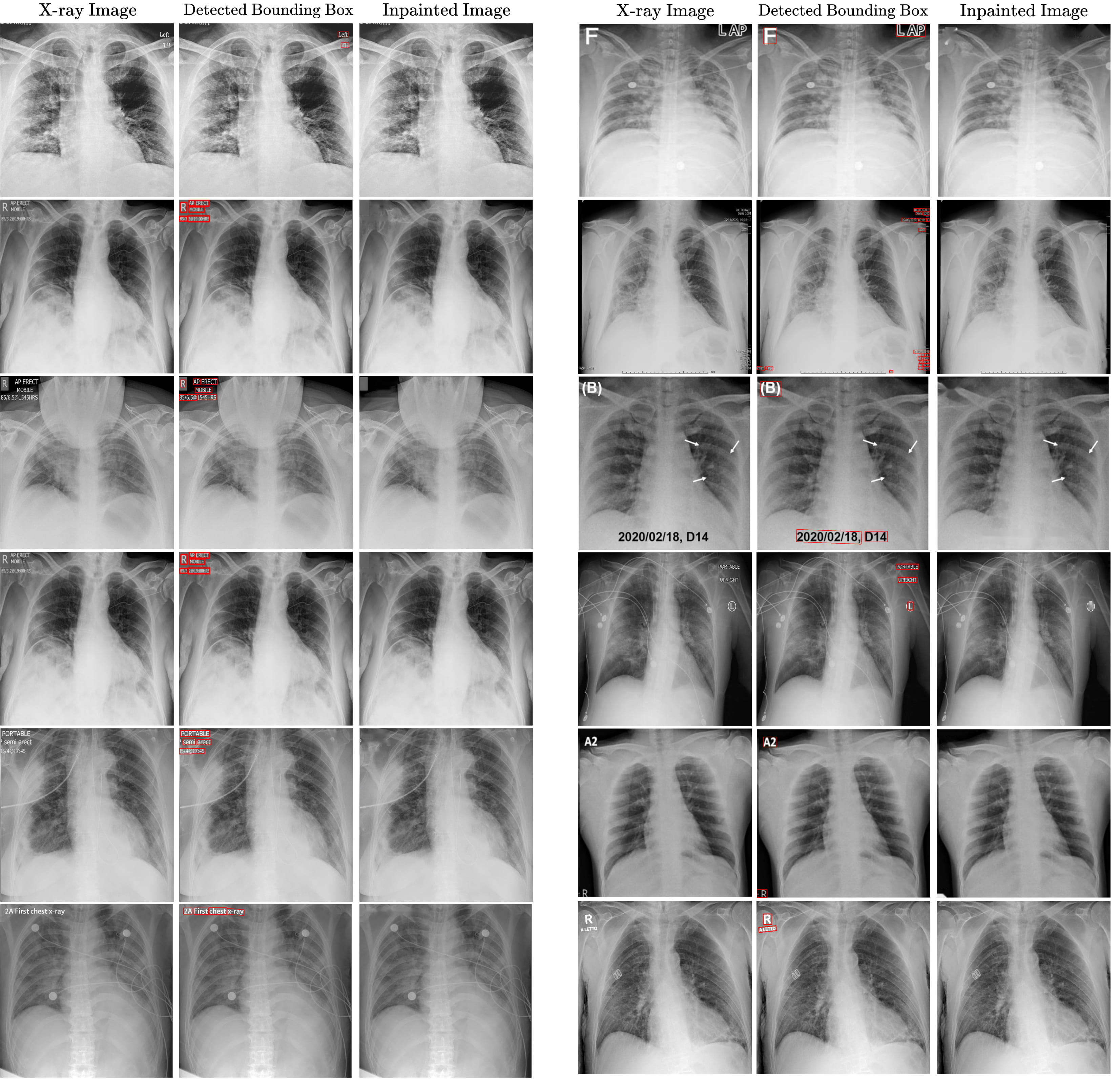}
}
\end{center}
% \vskip -0.15in
 \caption{Example of X-ray images from COVID-19 patients in the COVIDx dataset (left column), the detected bounding boxed using the CRAFT text detector pipeline (middle column), and the final inpainted images (COVIDx++ dataset).}
 \label{fig:inpaint}
\end{figure*}

%-------------------------
\textbf{D. \quad}\textbf{Data Cleaning for the Generation of The COVIDx++ dataset}: The first stage in cleansing the COVIDx dataset is to detect the text in the X-ray images. For this purpose, the CRAFT text detection method \cite{baek2019character} is used. We used the general pre-trained model available at the official code repository \footnote{\url{https://github.com/clovaai/CRAFT-pytorch}}. The text confidence threshold of 0.7 is used for the experiments. To inpaint inside the bounding boxes detected from the CRAFT, we used the built-in OpenCV inpainting function ({\texttt{cv2.inpaint(...)}}) with the algorithm by \cite{telea2004image} and inpaint radius of 10 pixels. Figure \ref{fig:inpaint} shows some samples from the COVID-19 patients in  the COVIDx dataset \cite{wang2020covidnet}, the corresponding detected bounding boxes detected by the CRAFT text detector, and the final inpainted images. \\

\begin{table}[htb]
% \vskip -0.15in
\caption{Running time comparison of iGOS++ against other visualization methods on ResNet50 at two different resolutions. The shown numbers are averaged over 5,000 images from the ImageNet validation set.}
\label{table:time}
% \vskip -0.05in
\begin{center}
\begin{small}
\begin{tabular}{c|c|c|c|c|c}
\toprule
{\bf Running} & {iGOS++} & {I-GOS} & {Mask} & {RISE} & {Gradient-} \\
{\bf Time (s)} & {} & {} & {} & {} & {Based} \\

\hline

28$\times$28    & {16.67} & {1.50}  & {14.66} & {-} & {\multirow{2}{*}{\bf $<$1}} \\
224$\times$224  & {12.25}  & {1.48} & {17.03} & {61.77} & {} \\

\bottomrule
\end{tabular}
\end{small}
\end{center}
\vskip -0.15in
\end{table}

%---------------------------------------------------------
\textbf{E. \quad}\textbf{Running Time.}
The results from running time comparison of our proposed method against the other perturbation-based methods, namely, I-GOS \cite{IGOS}, Mask \cite{ClassicMask}, and RISE \cite{2018RISE}, are presented in Table \ref{table:time} at two different resolutions. We followed the setting in \cite{IGOS} and an NVIDIA GeForce GTX 1080 Ti GPU along with an 8-core CPU is used for this experiment. The reported numbers are averaged over 5,000 images from the validation set of ImageNet. The gradient-based methods are faster to compute as they only need one forward and backward pass though the network --- generally speaking, this would take under one second per image to generate a heatmap. However, the resolution of their generated heatmaps is more restricted to the CNN architecture and not as flexible as in our method. They also are shown to perform poorer than our proposed method in both quantitative and qualitative evaluations. We improved the implementation code for \cite{IGOS}\footnote{\url{https://github.com/zhongangqi/IGOS}} and that is the reason the running time reported is faster than what the original paper reported. The main reason our method is slower than \cite{IGOS} is due to difference in calculating the step size using backtacking line search as explained in the methodology section. \\

%---------------------------------------------------------
%\textbf{Stability and Transferability}
%The same set of hyperparameters have been used throughout the experiments for both of the ImageNet and Fashion MNIST datasets as well for both of the ResNet50 and VGG19 models. We find our method robust against the choice of hyperparameters and resulting in slight difference in performance. Figure \ref{fig:hyp} shows visualization from our method on the Tiger Beetle image with three different sets of hyperparametrs $\lambda_1$ and $\lambda_2$ -- which have been shown to have the most influence on the final results. To find further details on the set of heyperparameters used as well as more figures, refer to the supplemenraty materials.

%------------------------
%\begin{figure}[hbt!]
%\begin{center}
%{ 
%\includegraphics[width=0.99\linewidth]{imgs/beet_hyp.pdf}
%}
%\end{center}
% \vskip -0.15in
% \caption{Visual comparison of our method (iGOS++) with different set of values for hyperparameter $\lambda_1$ and $\lambda_2$.}
% \label{fig:hyp}
%\end{figure}

\begin{figure*}[bt]
\begin{center}
{ 
\includegraphics[width=0.95\linewidth]{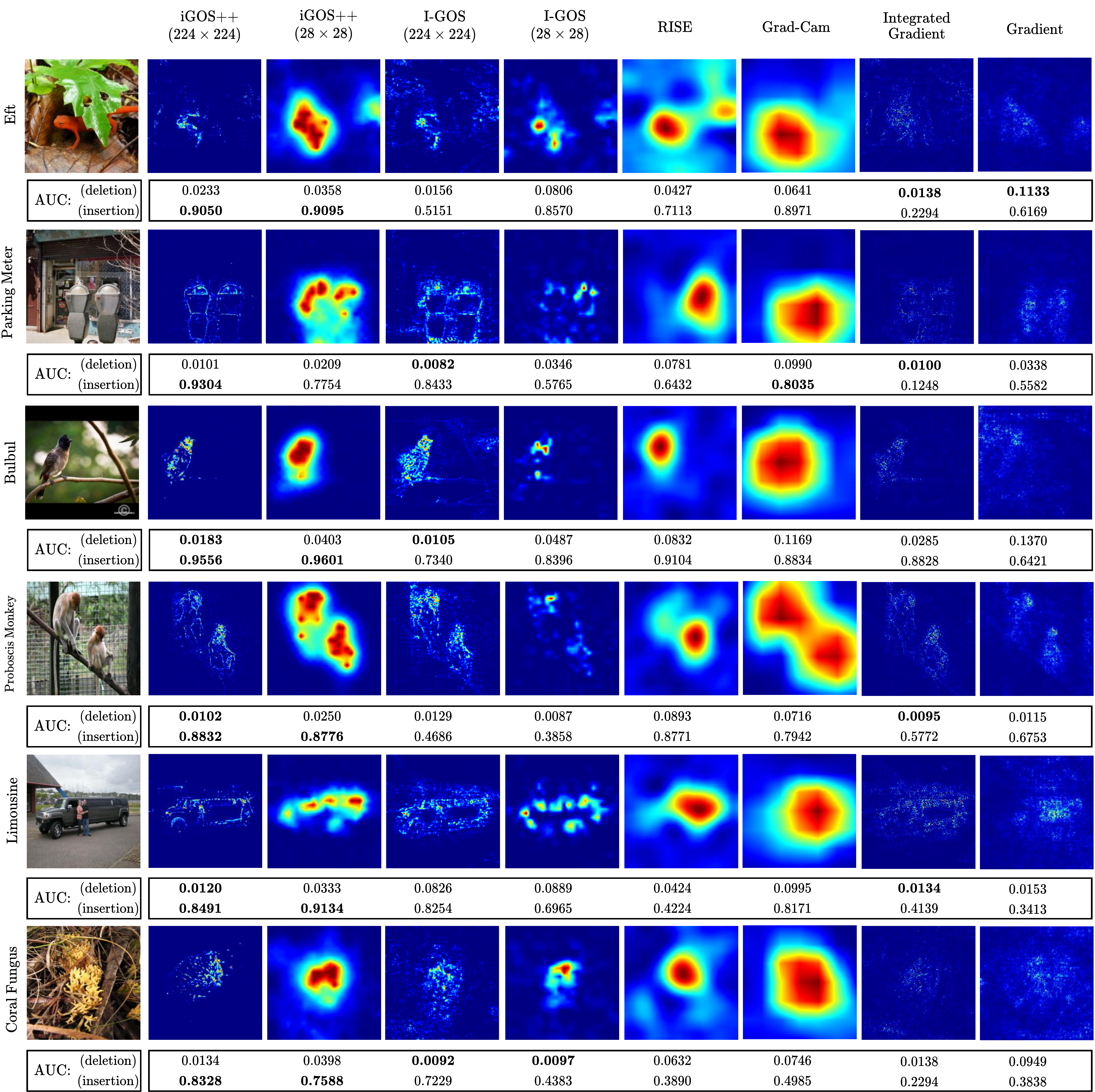}
}
\end{center}
\caption{\small iGOS++ visual explanations for different classes (written on the left) compared with other gradient and perturbation based visualization methods: I-GOS \cite{IGOS}, RISE \cite{2018RISE}, Grad-Cam \cite{Gradcam17}, Integrated-Gradient \cite{IntegratedGradient}, and Gradient \cite{SimonyanVZ13}. For the deletion AUC, lower is better. For the insertion AUC, higher is better. (Best viewed in color.) }
 \vskip -0.1in
 \label{fig:vis_supp}
\end{figure*}

%---------------------------------------------------------
\textbf{F. \quad}\textbf{iGOS++: Visual Comparisons.}
% \textbf{ImageNet.}
Figure. \ref{fig:vis_supp} compares the visual explanation from iGOS++ with other gradient-based (Grad-Cam \cite{Gradcam17}, Integrated-Gradient \cite{IntegratedGradient}, and Gradient \cite{SimonyanVZ13}) and perturbation based methods (I-GOS \cite{IGOS}, RISE \cite{2018RISE}). The images are selected from the ImageNet validation set, and the ResNet50 \cite{ResNet} is used as the classifier. The AUC for the deletion and insertion metrics are indicated under each visualization (for the deletion AUC, lower is better. For the insertion AUC, higher is better). For RISE visualizations, 8000 7$\times$7 random masks with $p=0.5$ are generated. The official implementations have been used for all the visualizations.\\
In Fig. \ref{fig:vis_supp}, it can be noted that Grad-Cam and RISE, due to the nature of their low-resolution masks (7$\times$7), highlight irrelevant regions in the image. In addition, Gradient method is calculated by the infinitesimal changes at the input image that change the prediction. To that end, as it can be seen in the figure, its visualizations are diffuse and not intuitive to human understanding. The same issue can be noted for Integrated-Gradient, however, it shows a good deletion score. Yet, Integrated-Gradient suffers in the insertion score. Similar to iGOS++, I-GOS has a flexibility in generating various resolution masks that can be chosen depending on the task at hand. Nevertheless, the visualizations from I-GOS are more scattered compared to the iGOS++ visualization (which are local on the object) and it suffers more on the insertion score. As an example, the visualization for the "BulBul" image (third row from the top) clearly underlines this issue.

%---------------------------

\begin{figure*}[bt]
\begin{center}
{ 
\includegraphics[width=0.825\linewidth]{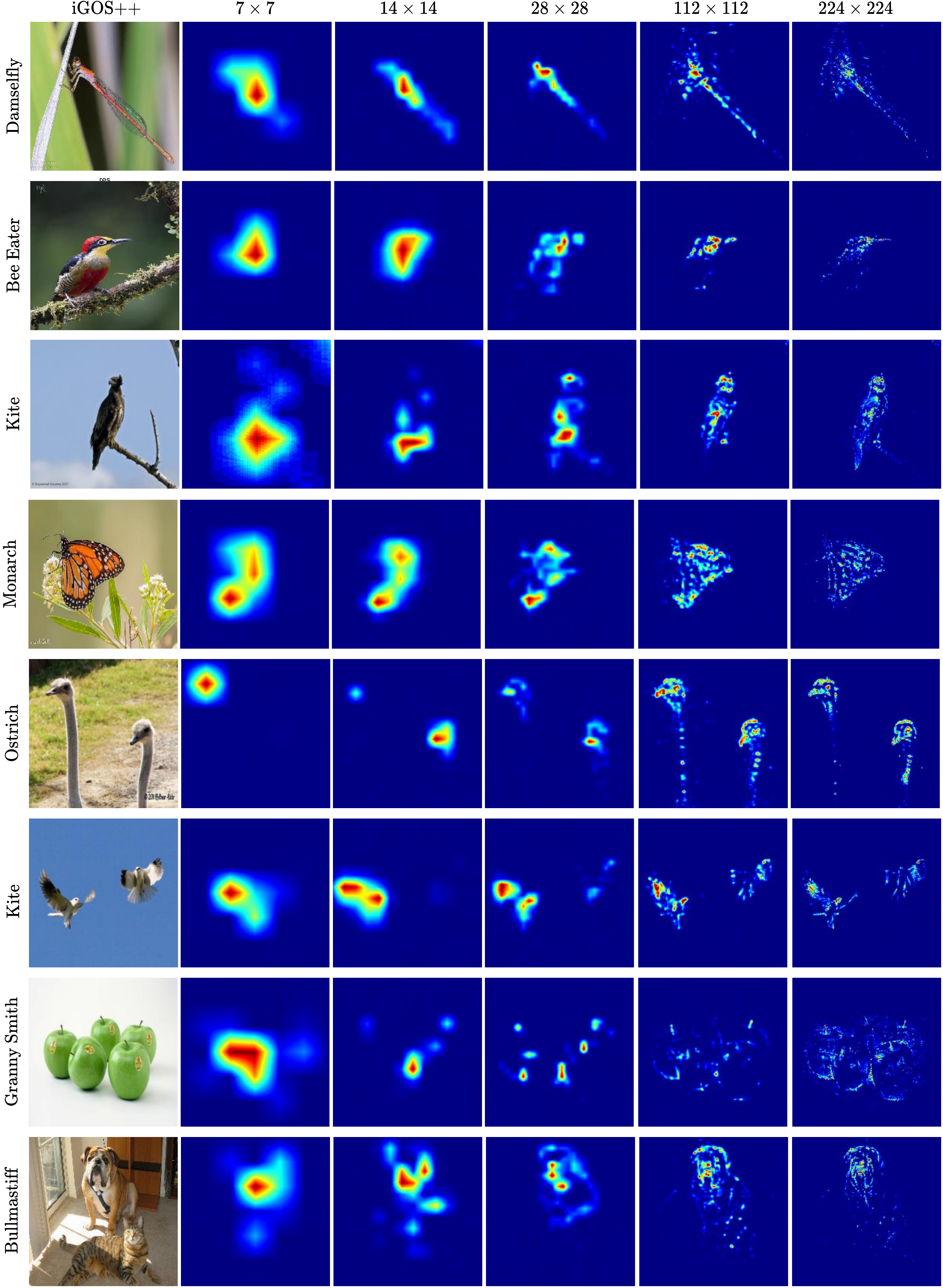}
}
\end{center}
\caption{\small iGOS++ visual explanations for different classes (written on the left) at different mask resolutions (written on the top). This figure shows the flexibility of our method in finding the most salient regions in the image from coarse and low resolution explanations (e.g. 7$\times$7) to refined and high resolution explanation (e.g. 224$\times$224). The last four rows also demonstrate examples where multiple objects are present in the image --- both from the same class, as in the "Ostrich", "Kite", and "Granny Smith" images, and from different classes, as in the "Bullmastiff" image. iGOS++ shows reliable explanation in both cases.}
 \vskip -0.1in
\label{fig:dif_res}
\end{figure*}

Figure \ref{fig:dif_res} compares the visualizations from iGOS++ at different resolutions. Our method has flexibility in the resolution of the generated mask and it can go from low-resolution and coarse visualizations (e.g. at 7$\times$7) to high-resolution and detailed explanations (e.g. 224$\times$224). It can be noted, when multiple objects (e.g. bottom four rows) are present in the image, low resolution masks perform poorer in locating them. This improves as the resolution of the explanations increases. For example, the "Granny Smith" image in the Fig. \ref{fig:dif_res} illustrates this. In addition, the visualization for the "Bullmastiff" image shows that iGOS++, unlike some visualization methods that perform (partial) image recovery \cite{adebayo2018sanity}, generates faithfull explanations when objects from different classes exist in the image.
%thin line for damselfly ...

%----------------------------

\begin{figure*}[ht!]
\begin{center}
% \subfigure[] { \label{fig:btv}
{
\includegraphics[width=0.8\linewidth]{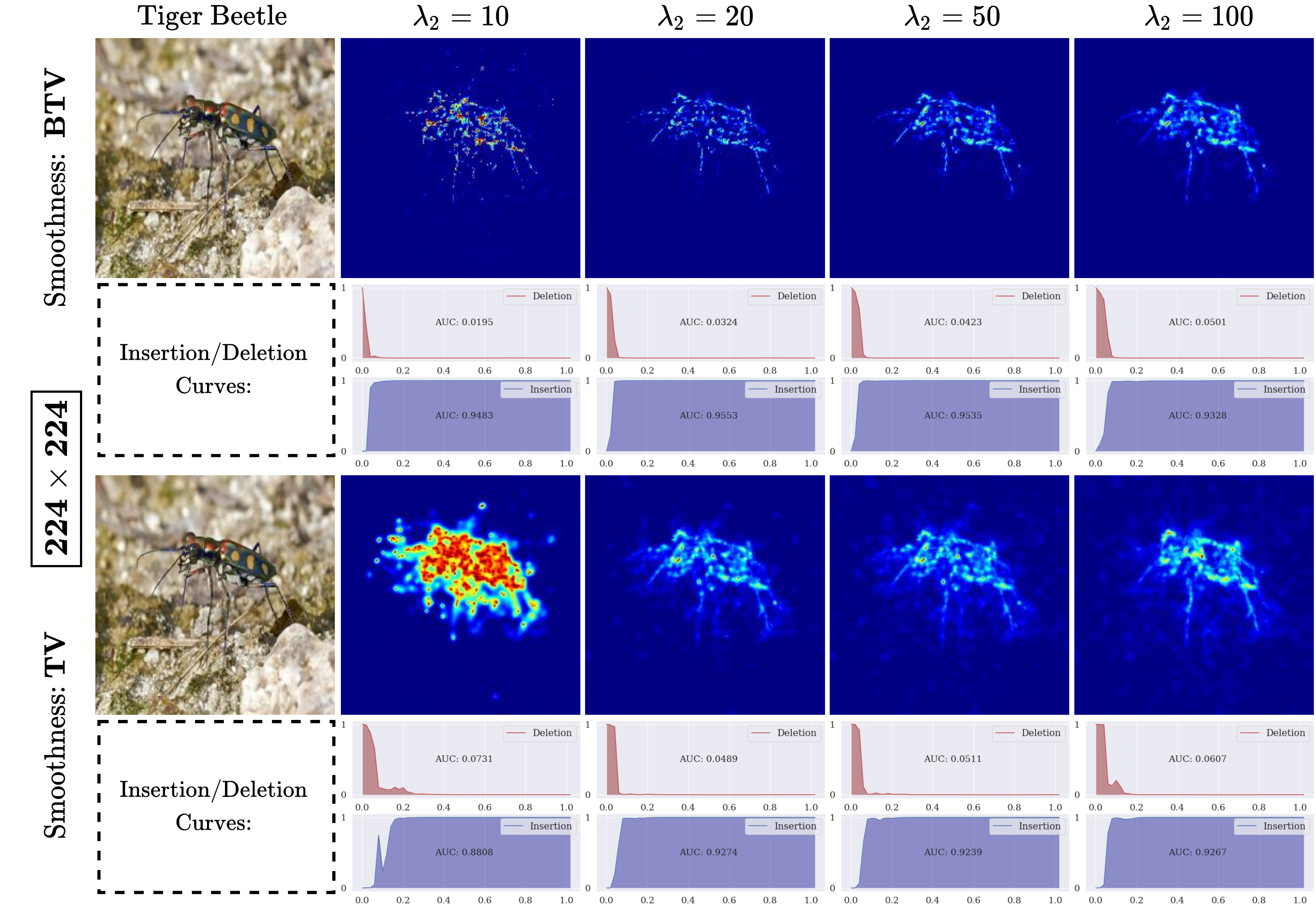}
}
% }
% \subfigure[] { \label{fig:block}
% \includegraphics[width=0.8\linewidth]{imgs/BTV_28_L1_1.pdf}
% }
\end{center}
% \vspace{-.15in}
\caption{\small Smoothness comparison of our proposed BTV term with the TV term \cite{ClassicMask} for the iGOS++ visualizations. Top row represents iGOS++ (with our proposed BTV term) while the bottom row shows when the TV term \cite{ClassicMask} is used instead. Each column shows different value for $\lambda_2$ hyperparameter (smoothness penalty). As it can be observed, when the $\lambda_2$ value increases, the generated masks become more smooth. In our experiments, we used $\lambda_2=20$ for all resolutions, models, and datasets. In addtion, it can be noted that BTV outperforms the TV as it generates a less scattered and local visualization which is more intuitive for human interpretation. }
 \label{fig:btv}
\end{figure*}

To visually compare the smoothness from our proposed BTV term with the TV term introduced in \cite{ClassicMask}, refer to Figure \ref{fig:btv} where the iGOS++ visualizations are shown for different smoothness losses as well as different values for their penalty constant $\lambda_2$. For this analysis, ResNet50 architecture is used. The generated mask for the "Tiger Beetle" are at 224$\times$224 resolution. It can be seen that when the BTV term is used (as in the original iGOS++ method), the visualizations are local to the object while for the TV term the visualizations are still scattered. For further details on the BTV term refer to the methodology section in the main paper. $\lambda_2=20$ is used in all of our experiments and different values in the Fig. \ref{fig:btv} are for highlighting the effect of the smoothness loss. 
%insertion/deletion curves.

%-----------------------------

\end{document}